\def\ARXIVFINAL{1}
\documentclass[11pt]{article}

\ifdefined\ARXIVFINAL
  \usepackage[final]{acl}
\else
  \usepackage[review]{acl}
\fi
\usepackage[T1]{fontenc}
\usepackage[utf8]{inputenc}
\usepackage{times}
\usepackage{latexsym}
\usepackage{kotex}
\usepackage{inconsolata}
\usepackage{booktabs}
\usepackage{graphicx}
\usepackage{float}
\usepackage{placeins}
\usepackage{array}
\usepackage{amsmath}
\usepackage{fvextra}
\usepackage{longtable}

\title{LLM-as-a-Judge Scores Are Unreliable Optimization Signals in
Closed-Loop Table Recognition}

\ifdefined\ARXIVFINAL
  \author{Donghwan Kim \\
    Aidentyx Inc., San Jose, CA, USA \\
    \texttt{david.kim@aidentyx.com}}
  \hypersetup{
    pdfauthor={Donghwan Kim}
  }
\else
  \author{Anonymous ACL submission}
\fi
\hypersetup{
  pdftitle={LLM-as-a-Judge Scores Are Unreliable Optimization Signals in Closed-Loop Table Recognition}
}

\begin{document}
\maketitle

\newcommand{\rid}[1]{\texttt{#1}}
\newcommand{\judgev}{\rid{judge\_v1}}
\newcommand{\calibv}{\rid{calibrated\_v2}}
\newcommand{\ablnf}{\rid{abl\_nofeedback}}
\newcommand{\cpnf}{\rid{cp\_nofeedback}}
\newcommand{\cpfb}{\rid{cp\_feedback}}
\newcommand{\crossnano}{\rid{cross\_gpt\_nano}}
\newcommand{\crossgpt}{\rid{cross\_gpt\_54}}
\newcommand{\crossopus}{\rid{cross\_claude\_opus}}
\newcommand{\policy}[1]{\mbox{#1}}

\begin{abstract}
LLM-as-a-judge is widely used to provide feedback and selection signals in
closed-loop regeneration, but this use remains insufficiently validated. We
study it in table recognition, where deterministic TEDS evaluation provides a
controlled testbed, using FinTabNet and OmniDocBench. Three findings emerge.
First, judge signals were weak on both datasets: scores frequently tied,
rankings were not reproducible, and no tested judge-score policy, whether selecting
among candidates or accepting revisions under a conservative score margin,
improved on the first output on both datasets. Iteration produced better
candidates, but the judge recovered them at most partially on one dataset
and not at all on the other. Second, severe losses occurred even without specific judge feedback,
supporting target-preservation failure under unconstrained regeneration as a
proximate mechanism. Third, a structure-preserving instruction reduced the
severe-loss rate, significantly on FinTabNet and directionally on
OmniDocBench, but produced no improvement, and in an exploratory
2$\times$2 analysis this protection was not stably observed when judge
feedback was retained. These results do not dispute the value of LLMs as evaluators, but show that
the tested reference-free judge signals were too weak and unstable to drive
candidate selection in this setup, and that
evaluation-style evidence alone was insufficient to establish closed-loop
optimization utility. Iterative refinement requires, at minimum, a
verification signal that deterministically detects structural change, rather
than judge scores alone.
\end{abstract}


\section{Introduction}
\label{sec:introduction}
LLM-as-a-judge uses an LLM to score output quality and has rapidly spread as a
way to reduce evaluation costs
\citep{zheng2023llmjudge,gu2026surveyllmasajudge}. Reported alignment with
human evaluation has expanded its role beyond evaluation. Closed-loop
pipelines now commonly use judge
scores for best-of-$n$ selection and inject the judge's error claims into the
prompt for repeated regeneration. This extension rests on an unvalidated
assumption: a judge able to evaluate an output can also guide its optimization.

We test this assumption empirically. The question is not whether a judge can
evaluate outputs, but whether its decisions are reliable enough to guide
optimization. The two capabilities have different requirements. Aggregate
alignment across many outputs may suffice for evaluation, whereas an
optimization signal must reproducibly rank subtly different candidates for the
same input. Recent work shows that aggregate judge alignment can overstate
utility for best-of-$n$ decisions \citep{landesberg2026bestofn}, and that judge
scores can diverge from independent quality signals during iterative
optimization with reference-free judges
\citep{pan2024rewardhacking,zhou2026selfplay}. We study a distinct
inference-time setting: closed-loop image-to-HTML regeneration. Without
exposing a reference inside the loop, we use a deterministic structural metric
to audit selection, feedback, and copy-preserving interventions at the
instance level. Unlike preference optimization or open-ended revision, we
isolate whether degradation arises from judge feedback itself or from
unconstrained regeneration.

Table recognition, which converts an image into an HTML table, provides an
unusually suitable setting for this test. TEDS is computed deterministically
against a fixed GT, allowing every judge decision to be compared against the
same benchmark objective. On two benchmarks comprising 476 FinTabNet tables
and 272 OmniDocBench tables, we run 8 iterations of judge-feedback-based
regeneration and observe judge scores and TEDS at every iteration. We compare
selection performance against a random baseline across multiple judge
configurations, two scoring formats (pointwise and pairwise), and three
tie-breaking rules (Section~3).

Our results are threefold. First, judge signals were weak on both datasets.
Scores from the judge that drove the loop tied on 37.5 to 49.1\% of
distinct-output candidate pairs, including, among pairs separated by at least
0.05 TEDS, 23.6 to 42.2\%. More granular scoring configurations also failed to
produce TEDS-aligned rankings, and rankings were not reproducible across
repeated scoring. One combination using the earliest tie-breaking rule beat random
selection on both datasets, but its advantage depended on the tie rule
and cannot be attributed to the judge scores alone.
No other combination of judge, scoring format, and tie-breaking rule robustly
beat random selection on both datasets. Iteration produced better candidates,
as indicated by significant oracle headroom, but the judge failed to recover
them. Guarded, margin-based acceptance did not change this conclusion: no
margin improved on the first output on both datasets, and on FinTabNet the
margins that avoided degradation accepted no revisions. Second, the loop's
net effect differed across datasets, ranging from
significant degradation on FinTabNet to a statistically neutral mean on
OmniDocBench. A stratified feedback audit identifies the pattern behind this
sign difference: it is consistent with variation in
the frequency of convention-sensitive structures. Third, we decompose a
proximate cause of degradation through a four-condition contrast. At the table
level, applying GT-inconsistent claims co-varied with breakage, but in the
paired comparison over the 99 tables shared with the original run, net
degradation remained the same without feedback content. A
structure-preservation constraint reduced the severe-loss rate without
feedback. This reduction was statistically significant in the full-sample
frozen-iter0 contrast on FinTabNet and was directionally consistent on
OmniDocBench. Thus, a proximate cause consistent with these observations is
target-preservation failure triggered by an unconstrained improvement
instruction, rather than feedback content.

Our contributions are as follows. (1) We show the absence of a robust selection signal across model
families, tiers, prompts, and scoring formats, find near-zero net
contribution from feedback content in a paired control, and examine
variation across generators.
(2) We decompose a proximate cause of degradation: severe losses occur without
specific judge feedback, and a
full-sample contrast branching from the same initial output shows that a
structure-preservation constraint significantly reduces the severe-loss rate
on FinTabNet and is directionally consistent on OmniDocBench. (3) We identify
boundary conditions and a mitigation. The net effect shows a monotonic
association with headroom, and a prompt-level structure-preservation constraint acts as a
conservative safeguard that reduces degradation or tail risk on both datasets.
The constraint leaves a safety-improvement trade-off, however, and exploratory results further suggest that retaining judge
feedback does not reliably preserve this safety benefit.
These conclusions are supported by a reproducible verification package
comprising independent metric validation, judge repeatability, an
independent-candidate control, artifact audits, a feedback-content contrast,
and a structure-preservation contrast. Our findings do not reject judges as
evaluators. They show that, in this setup, evaluation-style evidence for the
judges did not translate into closed-loop optimization utility, and they
identify the source of the gap.


\section{Related Work}
\label{sec:related-work}
\paragraph{LLM judges as evaluators.}
MT-Bench and Chatbot Arena popularized general-purpose LLM evaluation
\citep{zheng2023llmjudge}, and G-Eval reported dataset-level correlations
between GPT-4 scores and human evaluation for summarization and dialogue
generation \citep{liu2023geval}. Prometheus, trained with rubrics and reference
answers, and JudgeLM, trained to judge response pairs, demonstrate the
potential of customized judges \citep{kim2024prometheus,zhu2025judgelm}. In
the table domain, a GT-referenced judge has also been shown to score extraction
quality effectively \citep{horn2026beyond}. These positive results establish
that judges can be useful evaluators, although their validity varies with the
task and evaluation design \citep{bavaresco2025llmjudges}.

\paragraph{Reliability and decision validity.}
Dataset-level human alignment does not ensure correct within-input selection.
On LLMBar, even strong judges favored superficially better responses that
violated instructions \citep{zeng2024llmbar}. Order reversal reveals position
bias \citep{wang2024fairevaluators}, while repeated identical scoring can be
self-inconsistent \citep{haldar2025ratingroulette}. Factual errors, fabricated
citations, and format can shift both human and LLM judgments
\citep{chen2024judgementbias}. Trained judges degrade outside their training
evaluation methods \citep{empiricaljudge2024}, and a comparison across 20 NLP
tasks found that agreement varies by task, attribute, and annotator expertise
\citep{bavaresco2025llmjudges}. IF-RewardBench, which evaluates
rankings over preference graphs with multiple responses, likewise found
substantial gaps in constraint-violation detection and complex-instruction
ranking \citep{wen2026ifrewardbench}.

\paragraph{Selection and optimization signals, and reward hacking.}
A generation-verification gap can prevent a verifier from recovering an
available correct answer \citep{weaver2025}. In one-step best-of-$n$
selection, \citet{landesberg2026bestofn} found that a judge with reasonable
global correlation ($r=0.47$) recovered only 21\% of the oracle-available
improvement; ties and weak within-prompt rankings accounted for the gap, and
pairwise comparison raised recovery to 61\%. Evidence from iterative loops
is mixed \citep{gu2026surveyllmasajudge}: judge-generated rewards have
improved instruction following in iterative alignment training
\citep{yuan2024selfrewarding,wu2025metarewarding}, but model scores diverged
from human evaluation in self-evaluated essay revision
\citep{pan2024rewardhacking}, and self-play against a reference-free judge
raised judge pass rates while actual accuracy stagnated
\citep{zhou2026selfplay}, consistent with the broader risk of
over-optimizing a proxy \citep{gao2022overoptimization}. Verifiers trained
with external correctness supervision or step-level human labels can succeed
in mathematical best-of-N selection
\citep{cobbe2021verifiers,lightman2024verify,zhang2025generative}; our claim
is therefore limited to zero-shot reference-free judges, a boundary further
supported by evidence that cross-family verification is more useful on
average than self- or same-family verification \citep{lu2025verification}.

\paragraph{Feedback-driven self-correction.}
Automated correction is distinguished by the source and timing of its
feedback \citep{pan2024correction}. Self-Refine accumulates purely
self-generated feedback \citep{madaan2023selfrefine}, whereas Reflexion,
CRITIC, and MAF ground revision in environmental rewards, tool results, or
error-specific feedback modules
\citep{shinn2023reflexion,gou2024critic,nathani2023maf}, showing why
reliable external signals must be distinguished from self-feedback. Evidence
that prompted LLM feedback alone enables intrinsic self-correction remains
insufficient \citep{huang2023selfcorrect,kamoi2024selfcorrection}, and
repeated self-evaluation can amplify a preference for the model's own
outputs \citep{xu2024selfbias}. Multi-role critique has improved table
reasoning \citep{yu2025tablecritic}, but over-correction of correct answers
can offset the gains at high baseline accuracy
\citep{shaikh2026tableselfcorrect}, and even high-quality external feedback
grounded in GT may not be fully incorporated
\citep{jiang2025feedbackfriction}. In our setting, judge signal quality and
selection utility are major bottlenecks, and target preservation during
revision may be an independent one. Our loop uses reference-free
self-feedback from the same model without accumulating history and runs a
fixed number of zero-shot iterations (Appendix~\ref{app:G}).

\paragraph{Table recognition, GT conventions, and structural verification.}
PubTabNet introduced HTML tree-edit-based TEDS \citep{zhong2020teds}, GTE
constructed FinTabNet from financial documents \citep{zheng2021gte}, and
OmniDocBench spans diverse document types \citep{ouyang2025omnidoc}. GT,
however, is not a convention-independent representation: PubTables-1M
canonicalizes oversegmented annotations \citep{smock2022pubtables}, and
GriTS compares tables as 2D cell matrices rather than trees
\citep{smock2023grits}. Table-extraction work has used neighbor-guided
visual tools \citep{zhou2025ngtr} and combined symbolic checks with an LLM
\citep{mehrotra2026ten}. Grammar-constrained decoding reduces formatting
errors \citep{scholak2021picard,ugare2025syncode} but does not guarantee
deterministic execution or semantic agreement with image content, merge
structure, or GT conventions.

Our contribution is not the closed loop itself: prior work documents both
the potential and the failure of iterative evaluation and revision. We
instead audit, table by table, whether a reference-free judge improves
structural accuracy in image-based table recognition against fixed GT and
TEDS, jointly contrasting pointwise and pairwise selection, the presence or
absence of feedback, and unconstrained versus copy-preserving regeneration
from the same initial output. This design separates selection failure,
errors in feedback content, and target-preservation failure.


\section{Experimental Setup}
\label{sec:setup}

\subsection{Task and Iterative Loop}
\label{subsec:loop}

\begin{figure*}[t]
  \centering
  \includegraphics[width=\textwidth]{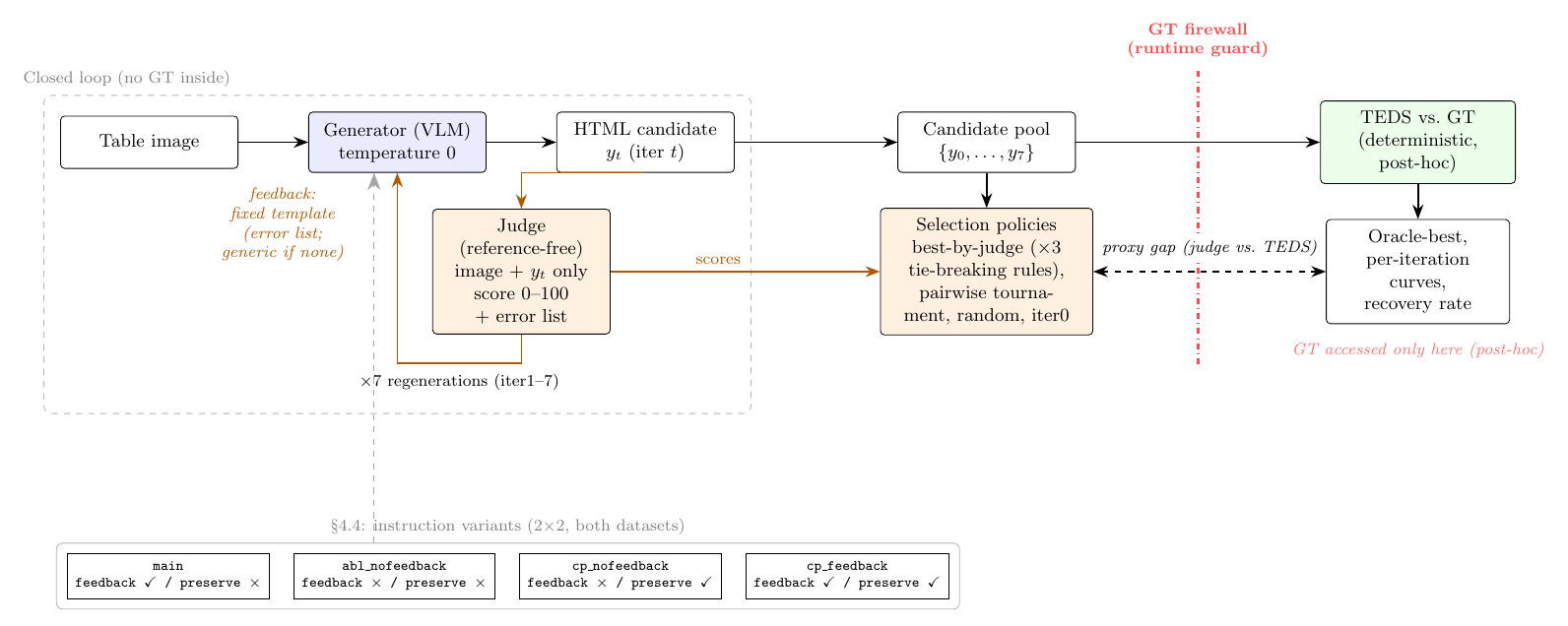}
  \caption{\textbf{Experimental framework.} The generator produces HTML from a
    table image, and the judge returns a score and an error list based only on
    the image and the immediately preceding output. This feedback is injected
    into the next regeneration step (7 iterations). GT and TEDS computation
    are used only for evaluation after the loop terminates and are not exposed
    during either generation or scoring (right of the dashed line). The
    instruction variants in the $2\times2$ contrast (Section~4.4) are injected
    into the generator prompt.}
  \label{fig:framework}
\end{figure*}

Our task is table recognition: given a table image, produce HTML
\texttt{table} markup that reproduces its structure and content. We use
Gemini 3.1 Flash Lite (version fixed at call time; Appendix~\ref{app:A}) as
the generator through OpenRouter. Both generation and scoring use temperature
0 and thinking level low. The output-token limits are 8192 for generation and
2048 for scoring.

The loop proceeds as follows (Figure~\ref{fig:framework}). At iteration 0
(iter0), the generator produces an initial HTML output from the table image
alone. At each subsequent iteration $t$ ($1\leq t\leq7$), (i) the judge takes
the table image and preceding output and returns a quality score from 0 to 100
together with an error list in JSON (enforced by the API's JSON schema), and
(ii) the generator produces a revision from the image, preceding output, and
judge feedback. The error list is inserted into a fixed template; when 0
errors are reported, the template instead supplies a generic improvement
instruction. We run a fixed total of 8 iterations, driven by \judgev{}
(Section~3.3).

Ground truth (GT) is never exposed at any generation or scoring step. At
manifest load, the runner does not load GT paths into memory, and it hooks file
opening so that any access to a GT directory fails immediately. Guard
installation and violations were recorded in audit logs, with 0 violations on
both datasets.

Failure handling, routing variability, temperature selection, and the fixed
iteration budget are detailed in Appendix~\ref{app:I}.

\subsection{Datasets}
\label{subsec:datasets}
\paragraph{FinTabNet.}
FinTabNet is a financial-table dataset introduced by GTE, which aligned PDF
and HTML representations from annual reports of U.S. S\&P 500 companies to
produce cell annotations \citep{zheng2021gte}. Stratified sampling over four
complexity strata yielded $n=476$ after iter0 failures. Financial-table
conventions such as section-label rows and currency-symbol columns are carried
into GT serialization. Provenance and sampling details are in
Appendix~\ref{app:I}.

\paragraph{OmniDocBench.}
OmniDocBench is a document-parsing benchmark spanning academic papers,
textbooks, newspapers, and handwritten notes \citep{ouyang2025omnidoc}.
Approximate stratification yielded $n=272$ after an iter0 failure. Because some
GT tables contain within-cell LaTeX expressions, directly
comparing absolute TEDS levels across the two datasets is inappropriate. We
compare only relative patterns under the same protocol.

The two datasets share the entire protocol, including the generator, judge,
prompts, iteration count, and temperature. Only the stratification criteria
differ approximately as described above.

\subsection{Judge Configurations}
\label{subsec:judges}
We evaluate selection and ranking using five judge configurations. All judges
are reference-free: they score only the table image and candidate HTML,
without GT. This design reflects the operating condition of the loop, because
the extraction task itself would be unnecessary if GT were available.

The configurations vary model family, tier, and score-production prompt. Cross
judges re-score stored outputs for counterfactual selection without rerunning
the loop. Evaluation sizes and selection procedures are in
Appendix~\ref{app:I}; prompts, model versions, and logs are in
Appendix~\ref{app:A}.

\subsection{Metrics, Selection Policies, and Statistics}
\label{subsec:metrics}
\paragraph{Metrics.}
TEDS~\citep{zhong2020teds} is widely used in table recognition. It represents
an HTML table as a tree and computes a tree-edit-distance similarity from 0 to
1, jointly evaluating structure and cell content. We also report S-TEDS, a
structure-only variant. GriTS is a complementary metric that compares tables
as 2D cell matrices and evaluates topology, location, and content in a common
framework \citep{smock2023grits}. GT is accessed only for post hoc evaluation.
Implementation validation and convention-dependent limitations are detailed in
Appendices~\ref{app:B}, \ref{app:C}, and \ref{app:I}.

\paragraph{Selection policies.}
We compare iter0, judge, pairwise, random, final, and oracle selection,
together with two guarded acceptance policies simulated post hoc on the
logged candidate stream: a margin rule that replaces the accepted output
(initially iter0) only when a later candidate's judge score exceeds it by at
least $\delta$, and a stop-on-clean rule that returns the first iteration
whose judge report contains no errors. Applied to candidates generated by
the original loop, these rules measure selection utility and do not alter
the generation trajectory.
Recovery is the mean improvement over iter0 divided by oracle headroom.
``Robustly beats'' requires cluster-bootstrap 95\% CIs supporting an advantage
over random on both datasets and under all tie-breaking rules. Full policy and
false-improvement definitions are in Appendix~\ref{app:I}.

\paragraph{Statistics.}
We report both Spearman correlation and Kendall tau-b because ties are
frequent, together with 95\% bootstrap CIs clustered by table. Paired
comparisons use the Wilcoxon signed-rank test. Given the nondeterminism of
judge scoring (Appendix~V2), selection-related point estimates are reported
together with the range of variation across repeated scoring.


\section{Results}
\label{sec:results}
\begingroup
\let\originalcaption\caption
\renewcommand{\caption}[1]{\originalcaption{\textbf{TEDS change relative to
iter0 and recovery rate by selection policy.} Recovery is the fraction of
attainable improvement (oracle $-$ iter0) recovered (Section~3.4).
\policy{best-by-judge} is reported under 3 tie-breaking rules, and the
``vs.\ random'' column compares against \policy{random-among-8} and gives the
cluster-bootstrap 95\% CI verdict. The pairwise protocol is in the appendix.
Bold marks the best executable policy for each dataset.}}
\begin{table*}[t]
  \centering
  \scriptsize
  \setlength{\tabcolsep}{3.2pt}
  \renewcommand{\arraystretch}{1.08}
  \caption{}
  \label{tab:selection-policies}
  \resizebox{\textwidth}{!}{%
  \begin{tabular}{llrrrrlrrrrl}
    \toprule
    & &
    \multicolumn{5}{c}{FinTabNet} &
    \multicolumn{5}{c}{OmniDocBench} \\
    \cmidrule(lr){3-7}\cmidrule(lr){8-12}
    Policy & Tie-breaking rule &
    $n$ & Mean $\Delta$ & Recovery & \multicolumn{2}{c}{vs.\ random (95\% CI)} &
    $n$ & Mean $\Delta$ & Recovery & \multicolumn{2}{c}{vs.\ random (95\% CI)} \\
    \midrule
    \policy{always-iter0} & \textemdash &
      476 & \textbf{0.0000} & \textbf{0.0\%} & \multicolumn{2}{l}{Baseline ($\Delta=0$)} &
      272 & 0.0000 & 0.0\% & \multicolumn{2}{l}{Baseline ($\Delta=0$)} \\
    \policy{best-by-\judgev} & earliest &
      476 & $-0.0062$ & $-33.2\%$ & \multicolumn{2}{l}{Better $[+0.0043,+0.0118]$} &
      272 & \textbf{$+0.0072$} & \textbf{27.7\%} & \multicolumn{2}{l}{Better $[+0.0003,+0.0092]$} \\
    \policy{best-by-\judgev} & latest &
      476 & $-0.0179$ & $-96.4\%$ & \multicolumn{2}{l}{Worse $[-0.0070,-0.0007]$} &
      272 & $+0.0052$ & 20.2\% & \multicolumn{2}{l}{No difference $[-0.0011,+0.0061]$} \\
    \policy{best-by-\judgev} & random (100 draws) &
      476 & $-0.0138$ & $-74.4\%$ & \multicolumn{2}{l}{No difference $[-0.0020,+0.0025]$} &
      272 & $+0.0056$ & 21.5\% & \multicolumn{2}{l}{Better $[+0.0001,+0.0060]$} \\
    \policy{best-by-\calibv} & earliest &
      476 & $-0.0030$ & $-16.2\%$ & \multicolumn{2}{l}{Better $[+0.0071,+0.0153]$} &
      \multicolumn{5}{c}{\textemdash} \\
    \policy{best-by-\crossnano} & earliest &
      143 & $-0.0061$ & $-30.7\%$ & \multicolumn{2}{l}{No difference $[-0.0034,+0.0075]$} &
      \multicolumn{5}{c}{\textemdash} \\
    \policy{best-by-\crossgpt} & earliest &
      75 & $-0.0102$ & $-43.4\%$ & \multicolumn{2}{l}{No difference $[-0.0103,+0.0142]$} &
      \multicolumn{5}{c}{\textemdash} \\
    \policy{best-by-\crossopus} & earliest &
      75 & $-0.0046$ & $-19.6\%$ & \multicolumn{2}{l}{No difference $[-0.0081,+0.0217]$} &
      \multicolumn{5}{c}{\textemdash} \\
    \policy{random-among-8} & \textemdash &
      476 & $-0.0143$ & $-76.8\%$ & \multicolumn{2}{l}{Reference} &
      272 & $+0.0026$ & 10.0\% & \multicolumn{2}{l}{Reference} \\
    \policy{always-final} & \textemdash &
      476 & $-0.0200$ & $-107.9\%$ & \multicolumn{2}{c}{\textemdash} &
      272 & $+0.0020$ & 7.8\% & \multicolumn{2}{c}{\textemdash} \\
    \policy{oracle-best} & \textemdash &
      476 & $+0.0186$ & 100.0\% & \multicolumn{2}{l}{Upper bound} &
      272 & $+0.0260$ & 100.0\% & \multicolumn{2}{l}{Upper bound} \\
    \bottomrule
  \end{tabular}%
  }
  \vspace{2pt}

  \begin{minipage}{0.99\textwidth}
    \footnotesize
    Bold marks the best executable policy for each dataset; \policy{oracle-best} is
    excluded because it requires ground truth. ``Better'', ``worse'', and
    ``no difference'' compare against \policy{random-among-8} using the clustered
    bootstrap interval shown.
  \end{minipage}
\end{table*}

\endgroup

Detailed judge diagnostics are reported in Appendix~\ref{app:I}.

\ifdefined\ARXIVFINAL
\begin{figure*}[t]
  \centering
  \includegraphics[width=\textwidth]{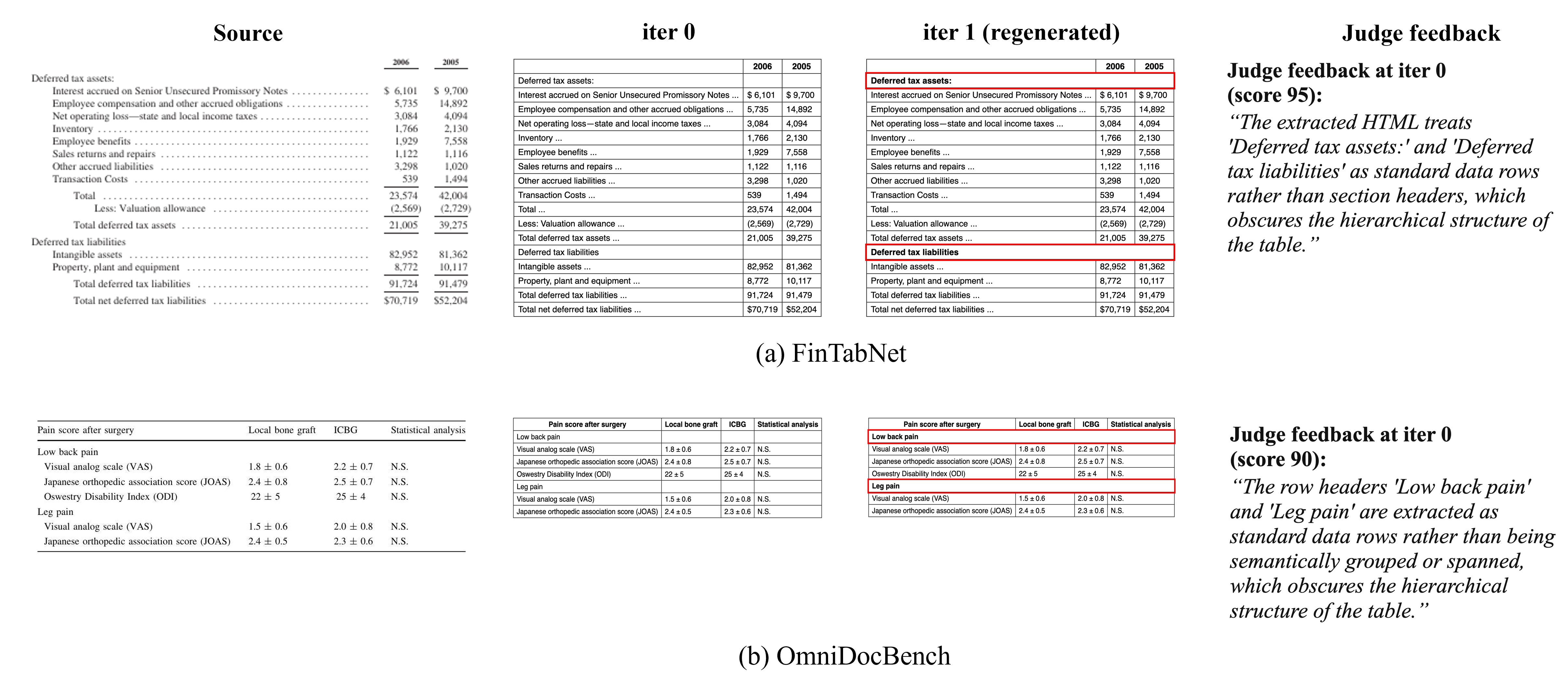}
  \caption{\textbf{The same type of breakage observed on both datasets.}
    Top (a), FinTabNet (TDG.2006); bottom (b), OmniDocBench (scihub). Each row,
    from left to right, shows the source image, the iter0 output, the iter1
    output regenerated using judge feedback (changed cells marked in red), and
    the verbatim judge feedback injected into that regeneration (iteration 0
    record, with score). In both examples, iter0 places a section label in a
    regular first-column cell according to the GT convention. The judge flags
    this as an error that obscures hierarchy, and the generator converts the
    label into a full-width merged header, departing from GT. The iter0-to-iter1
    difference is limited to the marked row, as verified by pixel comparison.
    A comparison with the GT structure is in the appendix.}
  \label{fig:breakage}
\end{figure*}
\fi

\subsection{Does the Loop Improve Quality? (RQ1)}
\label{subsec:rq1}
\ifdefined\ARXIVFINAL
\begin{figure*}[t]
  \centering
  \includegraphics[width=\textwidth]{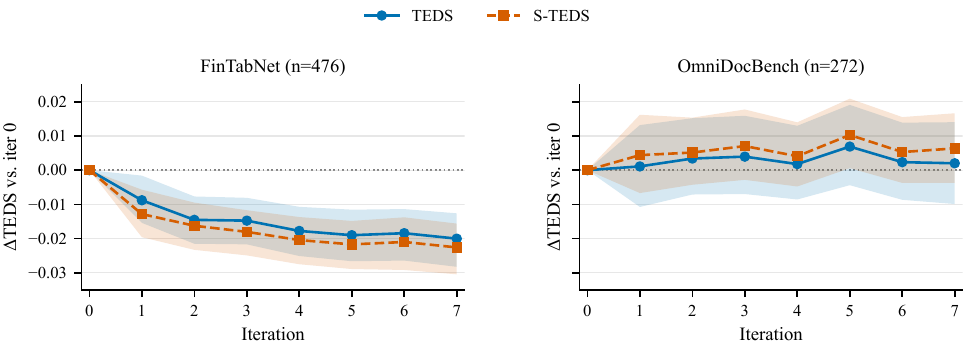}
  \caption{\textbf{Mean TEDS and S-TEDS by iteration on both datasets.}
  Values are changes relative to iter0. Shading gives table-cluster bootstrap
  95\% CIs.}
  \label{fig:iteration-curves}
\end{figure*}
\fi
The net effect of the iterative loop differs by dataset. On FinTabNet, the
loop degrades quality: the output after all 8 iterations (final) is on average
0.020 below the first output (iter0), with TEDS falling from 0.792 to 0.772
(Wilcoxon signed-rank $p<10^{-7}$). As
Figure~\ref{fig:iteration-curves} shows, the decline is gradual but begins
quickly: 44\% of the total drop occurs immediately after the first
regeneration, between iter0 and iter1. Because generation is
near-deterministic (Section~3.1), this first decline reflects the effects of
the elements newly introduced at regeneration: judge feedback, the preceding
output, and the improvement instruction. We separated these elements in a
control experiment. In a control loop that removed judge feedback content and
retained only conditioning on the preceding output and a generic improvement
instruction, net degradation remained the same size in the paired comparison
over the 99 tables completed in both runs (the net contribution of feedback
content was $+0.0001$, not significant; Appendix~\ref{app:B}). Thus, feedback
content is not the source of net degradation. Feedback does, however, increase
the amount of output change: without feedback, the proportion of stagnant
transitions whose consecutive iterations were unchanged rose from 54\% to
84\%. Section~4.4 analyzes the conditions under which regeneration produces
severe loss through a frozen-iter0 condition contrast. On OmniDocBench, in
contrast, the net effect of the loop is neutral (final $-$ iter0 = $+0.002$,
not significant).

However, ``neutral'' does not mean ``safe.'' When decomposed table by table
(Appendix~F), the loop substantially perturbs outputs on both datasets. On
FinTabNet, 41.6\% of tables worsen and only 23.9\% improve. On OmniDocBench,
25.7\% worsen and 20.6\% improve, but the upward contribution ($+0.020$) and
downward contribution ($-0.018$) almost exactly cancel, leaving the mean near
0. Among tables that changed, the magnitude of change is larger on
OmniDocBench than on FinTabNet (mean absolute change 0.082 vs.\ 0.066). Severe
breakage ($\Delta<-0.10$) occurs in 11.8\% and 4.4\% of tables, respectively,
and as Section~4.2 shows, the judge fails to detect it. In short, the loop
produces large table-level variation on both datasets; the dataset changes
only the mean sign of this variation.

Section~4.3 explains why the datasets diverge. The breakage type that dominates
the decline is the same on both datasets; what differs is the surface area on
which that type can occur.

\subsection{Can the Judge Select? (RQ2)}
\label{subsec:rq2}
Table~\ref{tab:selection-policies} compares the selection policies. On
FinTabNet, every judge policy has negative recovery: the output selected by the
judge is on average worse than the first output, iter0. The trivial
\policy{always-iter0} baseline beats every judge policy. Only one combination
significantly beats random selection, namely the self-judge combined with the
earliest-iteration rule, and this advantage depended on the tie-breaking rule. Because
quality tends to decline across iterations (Figure~\ref{fig:iteration-curves}),
the rule ``on a tie, choose the earlier iteration'' can win without using the
judge score. On FinTabNet, the policy was statistically indistinguishable
from a tie-block permutation null in which judge scores carry no information
(Appendix~\ref{app:I-selection}); its advantage therefore cannot be
attributed to the judge signal alone. On
OmniDocBench, recovery is positive (+20 to 28\%). Yet every per-candidate
judge policy misses more than 72\% of the attainable improvement, the best
guarded margin misses over half (Appendix~\ref{app:guarded}), and the gap to
the oracle is significant on both datasets ($p<10^{-29}$ and $p<10^{-16}$,
respectively).
Hereafter, ``robustly beats'' follows the definition in Section~3.4.

\paragraph{Guarded acceptance.}
Guarded rules did not recover the missed improvements
(Appendix~\ref{app:guarded}, Table~\ref{tab:guarded-grid}). On FinTabNet, no
margin improved mean TEDS over iter0: point estimates were negative for
$\delta\leq40$, and every margin with $\delta\geq60$ accepted zero revisions,
so the most cautious rules coincided with \policy{always-iter0}. On
OmniDocBench, the best margin ($\delta=20$, selected post hoc from the grid)
improved mean TEDS by $+0.0116$ over iter0 (95\% CI $[+0.0032, +0.0214]$;
recovery 44.5\%); hence no margin improved on iter0 on both datasets. The
$\delta=20$ rule beat
random selection on both datasets; on FinTabNet, however, this advantage
reflects rejecting nearly all revisions (12 of 476 tables changed) rather
than recovering better candidates. The stop-on-clean rule was worse than
iter0 on FinTabNet ($-0.0164$) and statistically neutral on OmniDocBench
($+0.0059$, CI includes zero).

The loop-driving judge tied on 49.1\% and 37.5\% of distinct-output
candidate pairs on FinTabNet and OmniDocBench, respectively, including
42.2\% and 23.6\% of pairs separated by at least 0.05 TEDS. Conversely,
cross judges with distinct-output tie rates as low as 6.4\% and 12.7\%
still showed near-zero or negative TEDS rank correlations, and rankings
were not reproducible across repeated scorings (Kendall's $W = 0.36$).

The failure was not merely low score resolution: rankings remained weak or
negative against TEDS and were not reproducible across repeated scoring,
models, prompts, or independent candidates. Pairwise comparison also failed to
beat random. This suggests that the bottleneck is the lack of a stable
within-instance selection signal in this candidate pool, not pointwise scoring
alone. We do not claim that pairwise comparison is generally invalid. Full
tie, null-model, and pairwise diagnostics are in Appendix~\ref{app:I}.

\subsection{Why Does It Fail? (RQ3)}
\label{subsec:rq3}

In a stratified audit, GT-inconsistent claims were concentrated among the
largest losses, while real claims were concentrated among improvements.
Addressing GT-inconsistent claims co-varied with breakage, but this
small-sample association should not be read as causal. Most conflicts favored
semantic HTML conventions over GT serialization, and S-TEDS confirmed
structural damage. Qualitatively similar convention-driven breakage was observed in
examples from both datasets. Breakage also
occurred with no specific claim, and net degradation remained unchanged
without feedback content. Thus, feedback can promote a shared model prior, but
unconstrained regeneration is sufficient for the observed damage. Full label
counts, confidence intervals, validation, and examples are in
Appendix~\ref{app:I}.

\begingroup
\let\originalcaption\caption
\renewcommand{\caption}[1]{\originalcaption{\textbf{Full-n frozen-iter0 B/C
contrast of severe-loss rates.} B is no-feedback unconstrained regeneration,
and C is no-feedback copy-preserving regeneration. Both conditions re-fork
from the same frozen iter0. Carry-forward is the primary analysis. The
difference is C$-$B in percentage points (pp), and the primary test is the
two-sided exact McNemar test. Because bootstrap CIs can be optimistic with
sparse events, they are reported as sensitivity evidence in
Appendix~\ref{app:B}.}}
\begin{table*}[t]
  \centering
  \scriptsize
  \setlength{\tabcolsep}{5.5pt}
  \renewcommand{\arraystretch}{1.12}
  \caption{}
  \label{tab:decomposition}
  \resizebox{\textwidth}{!}{%
  \begin{tabular}{lrrrrrr}
    \toprule
    Dataset & $n$ & B severe & C severe &
    C$-$B (pp) &
    \shortstack{B-only/C-only\\discordant pairs} &
    two-sided exact McNemar $p$ \\
    \midrule
    FinTabNet & 476 & 17/476 (3.6\%) & 4/476 (0.8\%) &
      $-2.7$ & 15/2 & 0.0023 \\
    OmniDocBench & 272 & 6/272 (2.2\%) & 2/272 (0.7\%) &
      $-1.5$ & 4/0 & 0.1250 \\
    \bottomrule
  \end{tabular}%
  }
  \vspace{2pt}

  \begin{minipage}{0.96\textwidth}
    \footnotesize
    B is no-feedback unconstrained regeneration and C is no-feedback
    copy-preserving regeneration from the same frozen iter0. Carry-forward is
    the primary failure handling rule. Because bootstrap intervals can be
    optimistic with few discordant events, the primary decision uses the
    two-sided exact McNemar test; bootstrap intervals and complete-case
    sensitivity are reported in Appendix~\ref{app:B}.
  \end{minipage}
\end{table*}

\endgroup

\ifdefined\ARXIVFINAL
\begin{figure*}[t]
  \centering
  \includegraphics[width=\textwidth]{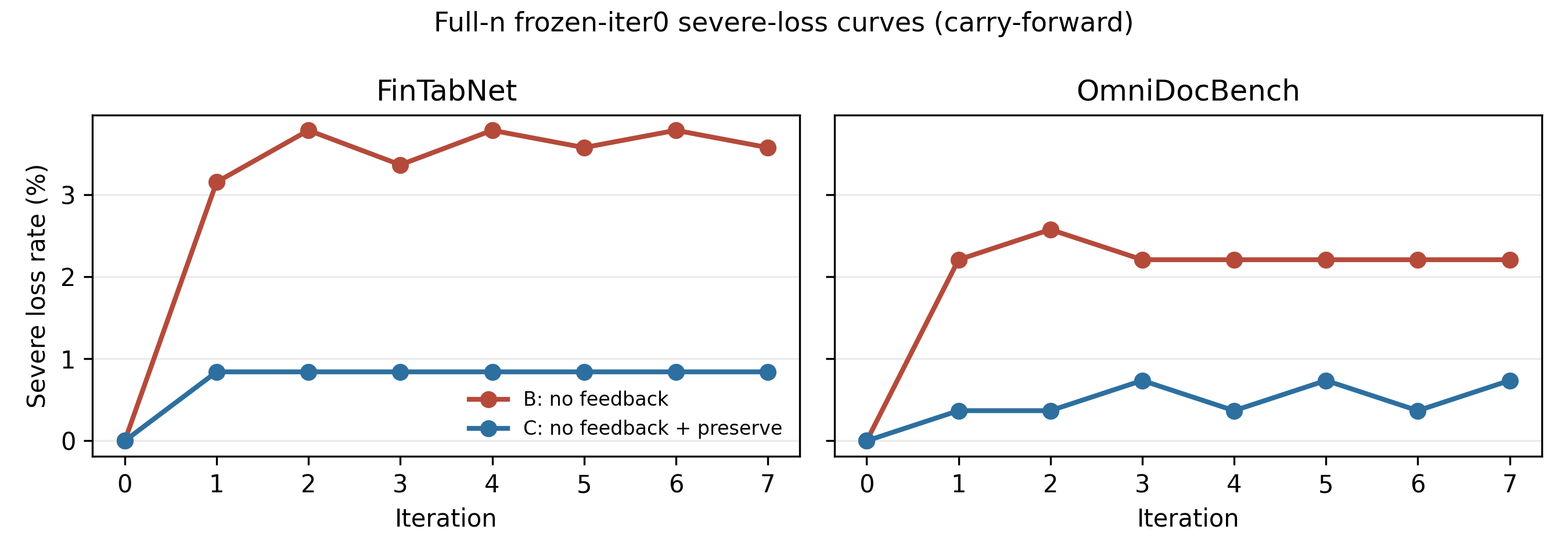}
  \caption{\textbf{Severe-loss rate by iteration in the full-n frozen-iter0
  B/C re-fork.} The final rate on FinTabNet fell from 3.6\% to 0.8\%
  (C$-$B = $-2.7$ pp, two-sided exact McNemar $p=0.0023$);
  OmniDocBench was directionally consistent, from 2.2\% to 0.7\%
  ($p=0.1250$).}
  \label{fig:conditions}
\end{figure*}
\fi

\subsection{What Causes the Degradation? (RQ4)}
\label{subsec:rq4}
The preceding sections showed that net degradation does not arise from
feedback content. To narrow its cause, we used a condition contrast crossing
two factors: the presence of judge feedback and the presence of a
structure-preserving instruction (Table~\ref{tab:decomposition};
Figure~\ref{fig:conditions}). The instruction directs the generator to retain
the preceding HTML unless clear visual evidence supports a change and not to
needlessly alter the numbers of rows and columns, merges, or placement of
empty cells. The primary analysis is a paired full-sample rerun in which both
conditions branch from the same stored iter0 output (hereafter, the
frozen-iter0 B/C contrast), fixed before inspecting the results. B supplies only a generic improvement instruction without feedback; C
adds the structure-preserving instruction under the same no-feedback
condition. Carry-forward is the primary failure handling rule, with
complete-case results reported as an appendix sensitivity analysis.

The primary result appears in the severe-loss tail. On FinTabNet, B had a
final severe-loss rate of 17/476 = 3.6\%, compared with 4/476 = 0.8\% for C.
The paired C$-$B difference was $-2.7$ pp, with 15/2 B-only/C-only discordant
pairs and a two-sided exact McNemar $p=0.0023$. OmniDocBench was directionally
consistent: B 6/272 = 2.2\%, C 2/272 = 0.7\%, C$-$B = $-1.5$ pp, 4/0
discordant pairs, and $p=0.1250$. We retained the severe-loss threshold of $\Delta$TEDS $< -0.10$
from the main-run analysis (Section 4.1), and the analysis plan and
primary endpoint were fixed before inspecting the frozen-iter0 contrast results. The effect was statistically significant
on FinTabNet and was directionally consistent on OmniDocBench.

Absolute run-level mean effects differed from the original experiment
(see Limitations); we therefore restrict inference to the paired
severe-loss contrast within the frozen-iter0 contrast.

We interpret mean changes as descriptive statistics only, not as equivalence
results. The mean $\Delta$ under B was $-0.0030$ on FinTabNet, with a
distinguishable location shift by the Wilcoxon signed-rank test
($p=0.0059$). It was $+0.0064$ on OmniDocBench, where the location shift was
not distinguishable ($p=0.1836$). These results do not show that a
structure-preservation constraint is an optimization method that produces
improvement. The narrower conclusion is that copy-preserving regeneration
without feedback reduces the severe-loss tail of unconstrained regeneration
without feedback.

This contrast shows that severe loss can occur without specific judge
feedback. Severe loss remains in B without feedback, while adding a
structure-preserving instruction under the same no-feedback condition in C
reduces that tail. The contrasts support target-preservation failure as a
proximate mechanism of the observed degradation: rather than a specific judge claim, an
improvement instruction given without a structure-preservation constraint
causes a structure already aligned with GT to be rewritten toward a learned
convention. The transition-level analysis of the main run also supports this
interpretation: 85.2\% of severe declines involved structural changes, and
large declines (greater than 0.05) occurred about 3 times as often in
transitions with structural changes as in content-only transitions (26.9\%
vs.\ 8.2\%; Appendix~\ref{app:H}). This is directionally consistent with the
breakage mechanism in Section~4.3, but the claim in this section is limited to
the severe-loss rate.

We do not elevate the interaction between feedback and the
structure-preservation constraint to a primary analysis in the main text. The
initial 100-table $2\times2$ experiment and complete-case sensitivity results
are retained in the appendix, while the carry-forward result from the
full-sample B/C analysis is primary. The conclusion is therefore conservative:
combining regeneration without feedback with a structure-preservation
constraint reduced the severe-loss rate. This effect was statistically
significant on FinTabNet and was directionally consistent on OmniDocBench. We
do not claim recovery of improvements or an increase in mean performance.


\section{Discussion}
\label{sec:discussion}
\paragraph{Evaluation ability and optimization utility are different
capabilities.}
Our results do not dispute the success of LLM-as-a-judge as an evaluator.
Results that establish dataset-level alignment with human evaluation or
successful GT-referenced table evaluation, including G-Eval, Prometheus, and
Horn and Keuper, can coexist with our findings
\citep{liu2023geval,kim2024prometheus,horn2026beyond}. The absence of a
reference is one important distinction between prior evaluation settings and
our closed-loop setting. Without GT, a judge must supply its own scoring
criterion. As our audit shows, that criterion is not the convention of the
target data, but the markup convention learned by the judge. Because the
absence of GT is a defining condition inside the loop, judge performance on
evaluation benchmarks does not guarantee performance as a loop signal.

\paragraph{Practical implications.}

The loop's net effect is difficult to predict without labeled validation, and
a neutral mean can conceal severe failures. Iter0 was the safest baseline.
No-feedback regeneration with a structure-preservation constraint reduced
severe loss, with statistical significance on FinTabNet and directional
consistency on OmniDocBench, but neither improved mean performance nor, in an exploratory
analysis, remained stably protective with feedback. Deterministic HTML guards can detect
structural changes missed by prompts. In this setting, judge scores alone should not gate deployment;
feedback requires claim verification, which still cannot prevent regeneration damage.
Details and the random-relative recovery comparison are in
Appendix~\ref{app:I}.


\section{Conclusion}
\label{sec:conclusion}
We analyzed an iterative table-recognition loop driven by an LLM judge using a
deterministic structural metric. The judge signal was weak or absent on both
datasets: scores frequently tied even across large TEDS gaps, more granular
configurations failed to produce TEDS-aligned rankings, and rankings outside
ties were not reproducible. One policy beat random only under an
earliest-iteration tie rule, consistent with exploiting the declining trend
rather than the judge signal; pairwise comparison did not help. The net effect ranged from significant degradation on FinTabNet to a
statistically neutral mean on OmniDocBench, consistent with a
difficult-to-predict surface on which markup preferences and data conventions
conflict. Better candidates
existed, but no judge-driven policy recovered them on both datasets.
Removing feedback left mean degradation unchanged on the 99 paired FinTabNet
tables, and copy-preserving no-feedback regeneration reduced severe loss in
the full-sample frozen-iter0 contrast (significant on FinTabNet, directionally
consistent on OmniDocBench; Section~4.4) without producing mean improvement.
A proximate cause consistent with these observations is
target-preservation failure triggered by an unconstrained improvement
instruction. Guarded, margin-based acceptance did not alter these
conclusions: no margin improved on the first output on both datasets, and
the margins that avoided degradation on FinTabNet accepted no revisions. In
this setup, evaluation-style evidence for the judges did not translate into
closed-loop optimization utility. Retaining the first output was the safest
baseline.


\section*{Limitations}
\label{sec:limitations}

\paragraph{Measurement.}
The TEDS implementation evaluates only the first table in an output, so quality
is underestimated for records in which the generator split a table (0.71\% of
all outputs). The conclusions are identical when these records are excluded or
corrected by merging the tables (Appendix~\ref{app:C}). The asymmetry by which
the judge sees the complete output while TEDS sees only the first table also
mechanically widens the judge-metric gap for these records. In addition, TEDS
depends on the GT serialization convention, so part of the proxy gap we report
is a convention conflict rather than a judge error. Oversegmentation and
canonicalization issues in table GT have also been documented in prior dataset
work \citep{smock2022pubtables,smock2023aligning}, and we did not apply an
alternative matrix-based metric such as GriTS throughout the experiment
\citep{smock2023grits}. However, S-TEDS also declined in the breakage examples,
so the structural damage itself cannot be reduced to a convention issue. We
did not perform human evaluation. Our claims concern optimization of the
benchmark objective, TEDS, rather than human-perceived table usability.

\paragraph{Scoring nondeterminism.}
Repeated scoring of the same input did not agree completely even at
temperature 0 (3-run exact agreement 68.6\%). Prior work reports that, in some
evaluation settings, averaging scores from sampling can align better with
human judgment than greedy decoding \citep{yamauchi2026design}. Our experiment
instead measured the selection stability of a single call as used in
deployment, not an average over repeated calls. Selection-related point
estimates such as recovery varied by tens of percentage points across
re-scoring runs. We therefore take the sign and policy ordering, rather than a
point estimate of recovery, as the unit of our claims. The use of more than one
serving backend is a possible cause, but we could not confirm it.

\paragraph{Design confounds and rule dependence.}
The independent-candidate control (Appendix~V4) removed feedback while also
changing generation temperature, so these two effects are not fully separated.
Self-judge recovery and its advantage over random depend strongly on the
tie-breaking rule, which exploits the decline in quality across iterations.
The \calibv{} prompt was selected against TEDS using 20 development tables.
Although the main evaluation is held out, we do not treat positive signals by
stratum as established claims. The single-elimination bracket in the pairwise
experiment may be sensitive to its seed, but repetition on 40 tables under 4
seeds produced an exact reproduction rate of only 20.5\% for the best
iteration. This reflects nondeterminism in the judge decisions, rather than
the bracket structure alone. The guarded-acceptance margin $\delta$ was
selected post hoc from a grid, so the OmniDocBench improvement at
$\delta=20$ is exploratory; the full grid is reported in
Appendix~\ref{app:guarded}. Guarded rules were simulated on the logged
candidate stream and do not capture how acceptance decisions would change
subsequent generations in a deployed loop.

\paragraph{Scope.}
Feedback labeling (Table~\ref{tab:feedback-labeling}) was performed by an LLM.
We release the rationale for each item to permit third-party verification and
validated the labels through a blinded author assessment
(Appendix~\ref{app:B}.4), but did not obtain an independent annotation from a
second annotator. We supplemented the generator axis with a trend check using 5
other families on 100 tables, holding the judge and protocol fixed and
changing only the generator (Appendix~E). Judge-score ties dominated for all
five generators. In contrast, the sign of the loop's net effect and the
utility of judge selection differed by generator. Both measures showed a
positive monotonic association with headroom, defined as oracle minus iter0
(Appendix~E; 6-point Spearman correlation 0.771 including main, $n=6$). The
generator with low baseline performance and large headroom showed a positive
net effect (approximately $+0.09$), and judge selection for that generator
exceeded random under all three tie-breaking rules on a single dataset of 100
tables. The generator with high baseline performance and limited headroom
showed a negative net effect (approximately $-0.06$). However, headroom cannot
be measured without ground truth, so we retain the conclusion that whether the
loop will help a particular generator or table cannot be predicted at
deployment time. This is trend evidence from 100 tables rather than an
established claim, and it does not determine whether the significant net
degradation in main generalizes across generators. Differences in iter0
baseline performance may confound comparisons of net effects across
generators. We use two datasets. Our conclusions are limited to the judge
configurations we tested and reference-free table-recognition loops; they do
not generalize to domains with verifiable answers, such as code with unit
tests.

\paragraph{Condition-asymmetric parsing failures.}
Because the four conditions are repeated measures on the same tables,
condition-asymmetric parsing failures require a repeated-measures test rather
than an ordinary chi-squared test; we used Cochran's Q. These failures may
confound interactions involving feedback and structure preservation, so those
interactions are reported as exploratory only. Detailed test and sensitivity
results are in Appendix~\ref{app:I}.

\paragraph{Non-replication of run-level mean effects.}
The mean net degradation in the unconstrained condition of the original
$2\times2$ experiment ($-0.0122$) did not recur in re-forking runs at later
times (unconstrained-condition means $+0.0000$ and $+0.0007$). The re-forking
was designed to detect paired contrasts between conditions and has limited
power to re-estimate run-level means. The use of more than one serving backend
is again a possible cause, but we could not confirm it. This non-replication is
directionally consistent with the implication in Section~5 that the net effect
of the loop is difficult to predict before deployment.


\begingroup
\sloppy
\hbadness=10000
\bibliography{references}
\endgroup

\clearpage
\appendix
\begingroup
\hbadness=10000
\vbadness=10000
\newcommand{\appasset}[2]{%
  \begin{figure*}[t]
    \centering
    \IfFileExists{#1}{%
      \includegraphics[width=0.92\textwidth]{#1}%
    }{%
      \fbox{\parbox[c][0.16\textheight][c]{0.88\textwidth}{%
        \centering Asset placeholder\\\texttt{\detokenize{#1}}%
      }}%
    }
    \caption{#2}
  \end{figure*}%
}
\newcommand{\appassetlabeled}[3]{%
  \begin{figure*}[t]
    \centering
    \IfFileExists{#1}{%
      \includegraphics[width=0.92\textwidth]{#1}%
    }{%
      \fbox{\parbox[c][0.16\textheight][c]{0.88\textwidth}{%
        \centering Asset placeholder\\\texttt{\detokenize{#1}}%
      }}%
    }
    \caption{#2}
    \label{#3}
  \end{figure*}%
}

\newcommand{\appendixassetroot}{..}
\IfFileExists{../scripts/looprunner/prompts.py}{}{%
  \renewcommand{\appendixassetroot}{../..}}

\section{Prompts and Model Versions}
\label{app:A}
\label{app:prompts}

This section preserves snapshots of the prompts used in the experiments and
the resolved model versions. The prompt text is not translated or rewritten;
it is included verbatim from the repository.

\subsection{Generation prompts}

\begin{Verbatim}
"""프롬프트 정의 (사양서 4·5절). 전 프롬프트 버전 식별자 포함.

러너는 이 모듈의 build_* 함수만 사용한다.
"""

from __future__ import annotations

# ------------------------------------------------------------------ judge

JUDGE_SYSTEM = """You are a meticulous evaluator of table extraction quality. You will be
shown the original image of a table and the raw HTML code extracted from
it. Judge the extracted HTML by considering how it would render as a
table. Do not evaluate stylistic HTML code quality; evaluate whether the
rendered table implied by the HTML preserves the content, structure, and
cell-to-header associations of the original image. Respond ONLY with a
single JSON object matching the required schema."""

_JUDGE_SCORING_GUIDE = """First, enumerate up to 5 of the most significant errors or ambiguities
you can find (empty list if none). Then assign an integer score from 0
to 100 using this scoring guide:
- 100: Perfect reproduction; all content, structure, merged cells, and
  header associations are correct.
- 90: Only trivial formatting differences; no ambiguity in
  cell-to-header mapping.
- 75: Mostly correct, but contains minor structural or content errors
  that do not affect most interpretations.
- 50: Partially correct, but important rows, columns, merged cells, or
  header associations are missing or ambiguous.
- 25: Major extraction failure; only small parts of the table are usable.
- 0: The output is unrelated, empty, or not a table.

Return ONLY this JSON object:
{
  "errors": [
    {"description": "<specific error>", "severity": "critical|major|minor"}
  ],
  "score": <integer 0-100>
}"""

JUDGE_USER_V1 = """Extracted HTML (raw code):
{candidate_html}

Evaluate the extracted HTML against the table shown in the image.
Judge semantic preservation of values, headers, and their associations,
as well as the table structure (rows, columns, merged cells, header
hierarchy).

Apply this key test: Could a reader who sees ONLY the rendered form of
the extracted HTML — without access to the original image — unambiguously
reconstruct every cell-to-header mapping and all content of the original
table? If not, consider the extraction as failed in that respect and
lower the score accordingly.

""" + _JUDGE_SCORING_GUIDE

# judge_v2 (파일럿 민감도 변형판): 오류 열거 문단 제거, anchor+점수만
JUDGE_USER_V2 = """Extracted HTML (raw code):
{candidate_html}

Evaluate the extracted HTML against the table shown in the image.
Judge semantic preservation of values, headers, and their associations,
as well as the table structure (rows, columns, merged cells, header
hierarchy).

Assign an integer score from 0 to 100 using this scoring guide:
- 100: Perfect reproduction; all content, structure, merged cells, and
  header associations are correct.
- 90: Only trivial formatting differences; no ambiguity in
  cell-to-header mapping.
- 75: Mostly correct, but contains minor structural or content errors
  that do not affect most interpretations.
- 50: Partially correct, but important rows, columns, merged cells, or
  header associations are missing or ambiguous.
- 25: Major extraction failure; only small parts of the table are usable.
- 0: The output is unrelated, empty, or not a table.

Return ONLY this JSON object:
{{
  "errors": [],
  "score": <integer 0-100>
}}"""

JUDGE_CALIBRATED_V2_SYSTEM = """You are a meticulous evaluator of table extraction quality, using a strict,
decomposed scoring rubric. You will be shown the original image of a
table and the raw HTML extracted from it. In past evaluations, scores
clustered too high (most extractions scored 90+) even when many cells
or structural elements were actually wrong. Assume errors exist unless
you can positively verify otherwise, and use the FULL 0-100 range —
do not default to high scores. Respond ONLY with a single JSON object
matching the required schema."""

JUDGE_CALIBRATED_V2_USER = """Extracted HTML (raw code):
{candidate_html}

Step 1 — Cell-level content accuracy: Compare every data cell (not
headers) in the image to the extracted HTML. Estimate
content_accuracy_pct: the percentage of data cells whose text content
is completely and exactly correct (integer 0-100).

Step 2 — Structural accuracy: Compare row/column layout, merged cells
(rowspan/colspan), and header hierarchy between the image and the HTML.
Estimate structure_accuracy_pct: the percentage of structural elements
(rows, columns, merges, header levels) that are correctly reproduced
(integer 0-100).

Step 3 — Enumerate up to 5 of the most significant errors (empty list
if none), each tagged with severity.

Calibration anchors (your percentages should behave like these):
- 100%: every cell/structural element verified correct.
- ~90%: 1-2 minor mismatches out of many elements.
- ~70%: a handful of cells/structural elements wrong, most correct.
- ~50%: roughly half right, half wrong or unverifiable.
- ~25%: only a small fraction correct.
- 0%: unrelated, empty, or not a table.

If you are unsure whether a cell or structural element is correct,
count it as INCORRECT (do not round up generously).

Return ONLY this JSON object:
{
  "content_accuracy_pct": <integer 0-100>,
  "structure_accuracy_pct": <integer 0-100>,
  "errors": [
    {"description": "<specific error>", "severity": "critical|major|minor"}
  ]
}"""

JUDGE_CALIBRATED_V2_SCHEMA = {
    "type": "object",
    "properties": {
        "content_accuracy_pct": {"type": "integer"},
        "structure_accuracy_pct": {"type": "integer"},
        "errors": {
            "type": "array",
            "items": {
                "type": "object",
                "properties": {
                    "description": {"type": "string"},
                    "severity": {"type": "string",
                                 "enum": ["critical", "major", "minor"]},
                },
                "required": ["description", "severity"],
                "additionalProperties": False,
            },
        },
    },
    "required": ["content_accuracy_pct", "structure_accuracy_pct", "errors"],
    "additionalProperties": False,
}


JUDGE_CALIBRATED_V3_SYSTEM = """You are a strict, meticulous evaluator of table extraction quality using
an explicit point-deduction rubric. You will be shown the original
image of a table and the raw HTML extracted from it. Start from the
assumption that the extraction is imperfect and look for every error —
do not award high scores by default. Respond ONLY with a single JSON
object matching the required schema."""

JUDGE_CALIBRATED_V3_USER = """Extracted HTML (raw code):
{candidate_html}

Carefully compare the extracted HTML against the table in the image,
cell by cell and row by row. List EVERY error you find (do not cap the
list), each with:
- "description": what is wrong
- "severity": "critical" (wrong/missing header-to-cell mapping, missing
  or extra row/column, broken merge) | "major" (incorrect cell content,
  wrong merge span) | "minor" (formatting-only difference, e.g. missing
  trailing punctuation)
- "count": how many cells/rows/columns this single error affects
  (integer >= 1)

Apply this key test before finalizing your list: could a reader
reconstruct the ENTIRE original table (all values, all headers, all
merges) using ONLY the rendered HTML? Any cell or association that
fails this test must be logged as at least a "major" error. If the
output is unrelated, empty, or not a table, log a single "critical"
error covering the whole table with a large count.

Return ONLY this JSON object:
{
  "errors": [
    {"description": "<specific error>", "severity": "critical|major|minor",
     "count": <integer >= 1>}
  ]
}"""

JUDGE_CALIBRATED_V3_SCHEMA = {
    "type": "object",
    "properties": {
        "errors": {
            "type": "array",
            "items": {
                "type": "object",
                "properties": {
                    "description": {"type": "string"},
                    "severity": {"type": "string",
                                 "enum": ["critical", "major", "minor"]},
                    "count": {"type": "integer"},
                },
                "required": ["description", "severity", "count"],
                "additionalProperties": False,
            },
        },
    },
    "required": ["errors"],
    "additionalProperties": False,
}

# 최종 score는 모델의 산술에 의존하지 않고 코드에서 결정적으로 재계산한다
# (postprocess.py의 recompute_calibrated_score 참조). v1/v2(sensitivity)는
# 모델이 직접 score를 반환하므로 그대로 사용.

JUDGE_JSON_REMINDER = """

REMINDER: Your previous response could not be parsed. Respond with ONLY a
single valid JSON object matching the required schema — no markdown, no
code fences, no explanation."""

JUDGE_SCHEMA_V1 = {
    "type": "object",
    "properties": {
        "errors": {
            "type": "array",
            "items": {
                "type": "object",
                "properties": {
                    "description": {"type": "string"},
                    "severity": {"type": "string",
                                 "enum": ["critical", "major", "minor"]},
                },
                "required": ["description", "severity"],
                "additionalProperties": False,
            },
        },
        "score": {"type": "integer"},
    },
    "required": ["errors", "score"],
    "additionalProperties": False,
}

# 버전별 (system, user_template, json_schema) 레지스트리.
# json_schema=None인 버전은 별도 스키마 강제 없이 프롬프트 지시만으로 JSON 유도.
JUDGE_VERSIONS = {
    "judge_v1": (JUDGE_SYSTEM, JUDGE_USER_V1, JUDGE_SCHEMA_V1),
    "judge_v2": (JUDGE_SYSTEM, JUDGE_USER_V2, JUDGE_SCHEMA_V1),
    "judge_calibrated_v2": (JUDGE_CALIBRATED_V2_SYSTEM, JUDGE_CALIBRATED_V2_USER,
                            JUDGE_CALIBRATED_V2_SCHEMA),
    "judge_calibrated_v3": (JUDGE_CALIBRATED_V3_SYSTEM, JUDGE_CALIBRATED_V3_USER,
                            JUDGE_CALIBRATED_V3_SCHEMA),
}


def build_judge_prompt(candidate_html: str, version: str = "judge_v1"):
    """(system, user_text, prompt_version) 반환. 이미지는 러너가 앞에 붙임."""
    system, tmpl, _ = JUDGE_VERSIONS[version]
    # 모든 v1류 템플릿은 JSON 예시에 중괄호가 있어 str.format 불가 → 직접 치환
    user = tmpl.replace("{candidate_html}", candidate_html)
    return system, user, version


def judge_schema_for(version: str) -> dict:
    return JUDGE_VERSIONS[version][2]


# ------------------------------------------------ pairwise judge (선택 전용)
# judge_v1의 평가 기준 서술을 최대한 재사용하고, 점수 산출 부분만 A/B 선택으로
# 교체한 버전. 스냅샷: scripts/judge_prompt_versions/judge_pairwise_v1.md

PAIRWISE_JUDGE_SYSTEM = """You are a meticulous evaluator of table extraction quality. You will be
shown the original image of a table and two candidate HTML extractions
of it, labeled A and B. Judge each candidate by considering how it would
render as a table. Do not evaluate stylistic HTML code quality; evaluate
whether the rendered table implied by the HTML preserves the content,
structure, and cell-to-header associations of the original image.
Respond ONLY with a single JSON object matching the required schema."""

PAIRWISE_JUDGE_USER_V1 = """Candidate A (raw HTML code):
{candidate_a}

Candidate B (raw HTML code):
{candidate_b}

Evaluate each candidate against the table shown in the image.
Judge semantic preservation of values, headers, and their associations,
as well as the table structure (rows, columns, merged cells, header
hierarchy).

Apply this key test to each candidate: Could a reader who sees ONLY the
rendered form of the extracted HTML — without access to the original
image — unambiguously reconstruct every cell-to-header mapping and all
content of the original table? A candidate that fails this test in more
places is the less faithful one.

Decide which candidate more faithfully reproduces the content,
structure, and cell-to-header associations of the original table. You
must pick exactly one winner, even if the two candidates are very close
or the difference is small.

Return ONLY this JSON object:
{
  "choice": "A" or "B"
}"""

PAIRWISE_SCHEMA_V1 = {
    "type": "object",
    "properties": {
        "choice": {"type": "string", "enum": ["A", "B"]},
    },
    "required": ["choice"],
    "additionalProperties": False,
}

PAIRWISE_VERSIONS = {
    "judge_pairwise_v1": (PAIRWISE_JUDGE_SYSTEM, PAIRWISE_JUDGE_USER_V1,
                          PAIRWISE_SCHEMA_V1),
}


def build_pairwise_judge_prompt(candidate_a: str, candidate_b: str,
                                version: str = "judge_pairwise_v1"):
    """(system, user_text, prompt_version) 반환. 이미지는 러너가 앞에 붙임."""
    system, tmpl, _ = PAIRWISE_VERSIONS[version]
    user = (tmpl.replace("{candidate_a}", candidate_a)
                .replace("{candidate_b}", candidate_b))
    return system, user, version


def pairwise_schema_for(version: str) -> dict:
    return PAIRWISE_VERSIONS[version][2]


# -------------------------------------------------------------- generator

GEN_SYSTEM = "You are an expert at converting table images into faithful HTML tables."

GEN_REQUIREMENTS = """Requirements:
- Reproduce ALL cell contents exactly as shown, including numbers,
  symbols, and formatting-relevant text.
- Reproduce the table structure faithfully: use rowspan/colspan for
  merged cells, and preserve header rows/hierarchy using <thead> and
  <th> where appropriate.
- If a cell contains a run of leader dots/periods used for visual
  alignment (e.g. "Item . . . . . . . 123"), collapse the run to a
  single ellipsis "..." rather than reproducing every dot. Never repeat
  any character or token in a loop.
- Output ONLY the HTML table, starting with <table> and ending with
  </table>. No markdown, no explanation, no code fences."""

GEN_V1 = f"""Convert this table image into a complete HTML table.

{GEN_REQUIREMENTS}"""

GEN_FEEDBACK_V1 = f"""Convert this table image into a complete HTML table.

{GEN_REQUIREMENTS}

A previous attempt is shown below, along with reviewer feedback.
Produce an improved version that fixes the identified issues while
preserving everything that was already correct.

Previous attempt:
{{previous_html}}

Reviewer feedback (issues found):
{{judge_errors_formatted}}

Output ONLY the corrected HTML table."""

GEN_GENERIC_V1 = f"""A previous attempt at converting this table image to HTML is shown
below. Carefully compare it against the table image and produce a
corrected HTML table. Preserve everything that is already correct.

{GEN_REQUIREMENTS}

Previous attempt:
{{previous_html}}

Output ONLY the corrected HTML table."""

# ------------------------------------------- copy-preserving 변형 (cp_*)
# novelty 재구축 실험: unconstrained regeneration이 손상 원인인지 분리.
# 보존 제약을 명시해 "명확한 시각 증거 없으면 이전 출력 유지"를 강제한다.

CP_PRESERVE_CONSTRAINTS = """Preservation constraints (IMPORTANT):
- Treat the previous HTML as the default answer: keep it unchanged unless
  you have clear visual evidence from the image that a specific change is
  required.
- Do NOT needlessly change the number of rows or columns, the
  rowspan/colspan structure, or the placement of empty cells.
- Only make minimal edits that fix a clear omission or clear
  misrecognition; preserve everything else exactly as it is.
- If you find nothing that clearly needs fixing, output the previous HTML
  verbatim."""

GEN_CP_NOFEEDBACK_V1 = f"""A previous attempt at converting this table image to HTML is shown
below. Carefully compare it against the table image.

{GEN_REQUIREMENTS}

{CP_PRESERVE_CONSTRAINTS}

Previous attempt:
{{previous_html}}

Output ONLY the HTML table."""

GEN_CP_FEEDBACK_V1 = f"""Convert this table image into a complete HTML table.

{GEN_REQUIREMENTS}

A previous attempt is shown below, along with reviewer feedback.
Produce an improved version that fixes the identified issues while
preserving everything that was already correct.

{CP_PRESERVE_CONSTRAINTS}

Previous attempt:
{{previous_html}}

Reviewer feedback (issues found):
{{judge_errors_formatted}}

Output ONLY the HTML table."""


def format_judge_errors(errors: list[dict]) -> str:
    """judge errors 배열 → "- [severity] description" 줄."""
    if not errors:
        return "- (no specific issues listed; improve overall fidelity)"
    return "\n".join(f"- [{e.get('severity', 'unknown')}] {e.get('description', '')}"
                     for e in errors)


def build_gen_prompt(iteration: int, previous_html: str | None,
                     judge_errors: list[dict] | None,
                     judge_parse_failed: bool = False,
                     rewritten_instructions: str | None = None,
                     preserve: bool = False):
    """(system, user_text, prompt_version) 반환. 이미지는 러너가 앞에 붙임.

    - iteration 0                → gen_v1
    - judge 파싱 실패 후 회차     → gen_generic_v1 (이전 HTML 유지, 피드백 없음)
    - B안(meta-grader) 재작성 시  → meta-grader가 만든 지시문으로 교체
    - preserve=True (cp_* 모드)  → 보존 제약 변형 (피드백 유무에 따라 분기)
    - 그 외                      → gen_feedback_v1
    """
    if iteration == 0:
        return GEN_SYSTEM, GEN_V1, "gen_v1"

    assert previous_html is not None, "iteration>=1에는 previous_html 필수"

    if preserve:
        if judge_parse_failed or judge_errors is None:
            # cp_nofeedback 경로 (cp_feedback의 judge 파싱 실패 폴백 포함)
            user = GEN_CP_NOFEEDBACK_V1.replace("{previous_html}", previous_html)
            return GEN_SYSTEM, user, "gen_cp_nofeedback_v1"
        user = (GEN_CP_FEEDBACK_V1
                .replace("{previous_html}", previous_html)
                .replace("{judge_errors_formatted}",
                         format_judge_errors(judge_errors)))
        return GEN_SYSTEM, user, "gen_cp_feedback_v1"

    if judge_parse_failed or judge_errors is None:
        user = GEN_GENERIC_V1.replace("{previous_html}", previous_html)
        return GEN_SYSTEM, user, "gen_generic_v1"

    if rewritten_instructions is not None:
        # B안: Requirements 이후 부분을 meta-grader가 재작성한 지시문으로 교체
        user = f"""Convert this table image into a complete HTML table.

{rewritten_instructions}

Previous attempt:
{previous_html}

Output ONLY the corrected HTML table."""
        return GEN_SYSTEM, user, "gen_meta_rewritten_v1"

    user = (GEN_FEEDBACK_V1
            .replace("{previous_html}", previous_html)
            .replace("{judge_errors_formatted}", format_judge_errors(judge_errors)))
    return GEN_SYSTEM, user, "gen_feedback_v1"


# ------------------------------------------------------------- meta-grader

META_SYSTEM = """You are a prompt engineer. Given reviewer feedback on a failed table
extraction, rewrite the extraction instructions to specifically prevent
the identified errors. Keep the instructions concise and imperative."""

META_USER_V1 = """The following errors were found in an HTML table extracted from an image:
{judge_errors_formatted}

Current instructions:
{current_instructions}

Rewrite the instructions to specifically address these errors. Return
ONLY the rewritten instructions, no explanation."""


def build_meta_prompt(judge_errors: list[dict], current_instructions: str):
    user = (META_USER_V1
            .replace("{judge_errors_formatted}", format_judge_errors(judge_errors))
            .replace("{current_instructions}", current_instructions))
    return META_SYSTEM, user, "meta_v1"
\end{Verbatim}

\subsection{Actual no-feedback wording for B}

The no-specific-issues fallback used in the actual B prompt was:
\begin{quote}
\texttt{- (no specific issues listed; improve overall fidelity)}
\end{quote}

\subsection{Pointwise \judgev{} prompt}

\paragraph{System prompt.}
\begin{Verbatim}
# judge_v1

이 파일은 `scripts/looprunner/prompts.py`의 스냅샷입니다. 실제 실행 시 소스는 prompts.py이며, 이 파일은 재현성/버전 추적을 위한 커밋용 기록입니다.

## Score 계산 방식

모델이 JSON의 `score` 필드로 직접 반환 (0-100).

## SYSTEM

```
You are a meticulous evaluator of table extraction quality. You will be
shown the original image of a table and the raw HTML code extracted from
it. Judge the extracted HTML by considering how it would render as a
table. Do not evaluate stylistic HTML code quality; evaluate whether the
rendered table implied by the HTML preserves the content, structure, and
cell-to-header associations of the original image. Respond ONLY with a
single JSON object matching the required schema.
```

## USER (템플릿, {candidate_html}은 추출된 HTML로 치환)

```
Extracted HTML (raw code):
{candidate_html}

Evaluate the extracted HTML against the table shown in the image.
Judge semantic preservation of values, headers, and their associations,
as well as the table structure (rows, columns, merged cells, header
hierarchy).

Apply this key test: Could a reader who sees ONLY the rendered form of
the extracted HTML — without access to the original image — unambiguously
reconstruct every cell-to-header mapping and all content of the original
table? If not, consider the extraction as failed in that respect and
lower the score accordingly.

First, enumerate up to 5 of the most significant errors or ambiguities
you can find (empty list if none). Then assign an integer score from 0
to 100 using this scoring guide:
- 100: Perfect reproduction; all content, structure, merged cells, and
  header associations are correct.
- 90: Only trivial formatting differences; no ambiguity in
  cell-to-header mapping.
- 75: Mostly correct, but contains minor structural or content errors
  that do not affect most interpretations.
- 50: Partially correct, but important rows, columns, merged cells, or
  header associations are missing or ambiguous.
- 25: Major extraction failure; only small parts of the table are usable.
- 0: The output is unrelated, empty, or not a table.

Return ONLY this JSON object:
{
  "errors": [
    {"description": "<specific error>", "severity": "critical|major|minor"}
  ],
  "score": <integer 0-100>
}
```

## JSON Schema

```json
{
  "type": "object",
  "properties": {
    "errors": {
      "type": "array",
      "items": {
        "type": "object",
        "properties": {
          "description": {
            "type": "string"
          },
          "severity": {
            "type": "string",
            "enum": [
              "critical",
              "major",
              "minor"
            ]
          }
        },
        "required": [
          "description",
          "severity"
        ],
        "additionalProperties": false
      }
    },
    "score": {
      "type": "integer"
    }
  },
  "required": [
    "errors",
    "score"
  ],
  "additionalProperties": false
}
```
\end{Verbatim}
\paragraph{User prompt template.}
\begin{Verbatim}
# judge_v1

이 파일은 `scripts/looprunner/prompts.py`의 스냅샷입니다. 실제 실행 시 소스는 prompts.py이며, 이 파일은 재현성/버전 추적을 위한 커밋용 기록입니다.

## Score 계산 방식

모델이 JSON의 `score` 필드로 직접 반환 (0-100).

## SYSTEM

```
You are a meticulous evaluator of table extraction quality. You will be
shown the original image of a table and the raw HTML code extracted from
it. Judge the extracted HTML by considering how it would render as a
table. Do not evaluate stylistic HTML code quality; evaluate whether the
rendered table implied by the HTML preserves the content, structure, and
cell-to-header associations of the original image. Respond ONLY with a
single JSON object matching the required schema.
```

## USER (템플릿, {candidate_html}은 추출된 HTML로 치환)

```
Extracted HTML (raw code):
{candidate_html}

Evaluate the extracted HTML against the table shown in the image.
Judge semantic preservation of values, headers, and their associations,
as well as the table structure (rows, columns, merged cells, header
hierarchy).

Apply this key test: Could a reader who sees ONLY the rendered form of
the extracted HTML — without access to the original image — unambiguously
reconstruct every cell-to-header mapping and all content of the original
table? If not, consider the extraction as failed in that respect and
lower the score accordingly.

First, enumerate up to 5 of the most significant errors or ambiguities
you can find (empty list if none). Then assign an integer score from 0
to 100 using this scoring guide:
- 100: Perfect reproduction; all content, structure, merged cells, and
  header associations are correct.
- 90: Only trivial formatting differences; no ambiguity in
  cell-to-header mapping.
- 75: Mostly correct, but contains minor structural or content errors
  that do not affect most interpretations.
- 50: Partially correct, but important rows, columns, merged cells, or
  header associations are missing or ambiguous.
- 25: Major extraction failure; only small parts of the table are usable.
- 0: The output is unrelated, empty, or not a table.

Return ONLY this JSON object:
{
  "errors": [
    {"description": "<specific error>", "severity": "critical|major|minor"}
  ],
  "score": <integer 0-100>
}
```

## JSON Schema

```json
{
  "type": "object",
  "properties": {
    "errors": {
      "type": "array",
      "items": {
        "type": "object",
        "properties": {
          "description": {
            "type": "string"
          },
          "severity": {
            "type": "string",
            "enum": [
              "critical",
              "major",
              "minor"
            ]
          }
        },
        "required": [
          "description",
          "severity"
        ],
        "additionalProperties": false
      }
    },
    "score": {
      "type": "integer"
    }
  },
  "required": [
    "errors",
    "score"
  ],
  "additionalProperties": false
}
```
\end{Verbatim}
\paragraph{JSON schema.}
\begin{Verbatim}
# judge_v1

이 파일은 `scripts/looprunner/prompts.py`의 스냅샷입니다. 실제 실행 시 소스는 prompts.py이며, 이 파일은 재현성/버전 추적을 위한 커밋용 기록입니다.

## Score 계산 방식

모델이 JSON의 `score` 필드로 직접 반환 (0-100).

## SYSTEM

```
You are a meticulous evaluator of table extraction quality. You will be
shown the original image of a table and the raw HTML code extracted from
it. Judge the extracted HTML by considering how it would render as a
table. Do not evaluate stylistic HTML code quality; evaluate whether the
rendered table implied by the HTML preserves the content, structure, and
cell-to-header associations of the original image. Respond ONLY with a
single JSON object matching the required schema.
```

## USER (템플릿, {candidate_html}은 추출된 HTML로 치환)

```
Extracted HTML (raw code):
{candidate_html}

Evaluate the extracted HTML against the table shown in the image.
Judge semantic preservation of values, headers, and their associations,
as well as the table structure (rows, columns, merged cells, header
hierarchy).

Apply this key test: Could a reader who sees ONLY the rendered form of
the extracted HTML — without access to the original image — unambiguously
reconstruct every cell-to-header mapping and all content of the original
table? If not, consider the extraction as failed in that respect and
lower the score accordingly.

First, enumerate up to 5 of the most significant errors or ambiguities
you can find (empty list if none). Then assign an integer score from 0
to 100 using this scoring guide:
- 100: Perfect reproduction; all content, structure, merged cells, and
  header associations are correct.
- 90: Only trivial formatting differences; no ambiguity in
  cell-to-header mapping.
- 75: Mostly correct, but contains minor structural or content errors
  that do not affect most interpretations.
- 50: Partially correct, but important rows, columns, merged cells, or
  header associations are missing or ambiguous.
- 25: Major extraction failure; only small parts of the table are usable.
- 0: The output is unrelated, empty, or not a table.

Return ONLY this JSON object:
{
  "errors": [
    {"description": "<specific error>", "severity": "critical|major|minor"}
  ],
  "score": <integer 0-100>
}
```

## JSON Schema

```json
{
  "type": "object",
  "properties": {
    "errors": {
      "type": "array",
      "items": {
        "type": "object",
        "properties": {
          "description": {
            "type": "string"
          },
          "severity": {
            "type": "string",
            "enum": [
              "critical",
              "major",
              "minor"
            ]
          }
        },
        "required": [
          "description",
          "severity"
        ],
        "additionalProperties": false
      }
    },
    "score": {
      "type": "integer"
    }
  },
  "required": [
    "errors",
    "score"
  ],
  "additionalProperties": false
}
```
\end{Verbatim}

\subsection{\calibv{} pointwise judge prompt}

\paragraph{System prompt.}
\begin{Verbatim}
# judge_calibrated_v2

이 파일은 `scripts/looprunner/prompts.py`의 스냅샷입니다. 실제 실행 시 소스는 prompts.py이며, 이 파일은 재현성/버전 추적을 위한 커밋용 기록입니다.

## Score 계산 방식

모델은 score를 반환하지 않는다. 코드가 결정적으로 재계산:
  score = round(0.5 * content_accuracy_pct + 0.5 * structure_accuracy_pct)
(postprocess.recompute_calibrated_score 참조)

## SYSTEM

```
You are a meticulous evaluator of table extraction quality, using a strict,
decomposed scoring rubric. You will be shown the original image of a
table and the raw HTML extracted from it. In past evaluations, scores
clustered too high (most extractions scored 90+) even when many cells
or structural elements were actually wrong. Assume errors exist unless
you can positively verify otherwise, and use the FULL 0-100 range —
do not default to high scores. Respond ONLY with a single JSON object
matching the required schema.
```

## USER (템플릿, {candidate_html}은 추출된 HTML로 치환)

```
Extracted HTML (raw code):
{candidate_html}

Step 1 — Cell-level content accuracy: Compare every data cell (not
headers) in the image to the extracted HTML. Estimate
content_accuracy_pct: the percentage of data cells whose text content
is completely and exactly correct (integer 0-100).

Step 2 — Structural accuracy: Compare row/column layout, merged cells
(rowspan/colspan), and header hierarchy between the image and the HTML.
Estimate structure_accuracy_pct: the percentage of structural elements
(rows, columns, merges, header levels) that are correctly reproduced
(integer 0-100).

Step 3 — Enumerate up to 5 of the most significant errors (empty list
if none), each tagged with severity.

Calibration anchors (your percentages should behave like these):
- 100%: every cell/structural element verified correct.
- ~90%: 1-2 minor mismatches out of many elements.
- ~70%: a handful of cells/structural elements wrong, most correct.
- ~50%: roughly half right, half wrong or unverifiable.
- ~25%: only a small fraction correct.
- 0%: unrelated, empty, or not a table.

If you are unsure whether a cell or structural element is correct,
count it as INCORRECT (do not round up generously).

Return ONLY this JSON object:
{
  "content_accuracy_pct": <integer 0-100>,
  "structure_accuracy_pct": <integer 0-100>,
  "errors": [
    {"description": "<specific error>", "severity": "critical|major|minor"}
  ]
}
```

## JSON Schema

```json
{
  "type": "object",
  "properties": {
    "content_accuracy_pct": {
      "type": "integer"
    },
    "structure_accuracy_pct": {
      "type": "integer"
    },
    "errors": {
      "type": "array",
      "items": {
        "type": "object",
        "properties": {
          "description": {
            "type": "string"
          },
          "severity": {
            "type": "string",
            "enum": [
              "critical",
              "major",
              "minor"
            ]
          }
        },
        "required": [
          "description",
          "severity"
        ],
        "additionalProperties": false
      }
    }
  },
  "required": [
    "content_accuracy_pct",
    "structure_accuracy_pct",
    "errors"
  ],
  "additionalProperties": false
}
```
\end{Verbatim}
\paragraph{User prompt template.}
\begin{Verbatim}
# judge_calibrated_v2

이 파일은 `scripts/looprunner/prompts.py`의 스냅샷입니다. 실제 실행 시 소스는 prompts.py이며, 이 파일은 재현성/버전 추적을 위한 커밋용 기록입니다.

## Score 계산 방식

모델은 score를 반환하지 않는다. 코드가 결정적으로 재계산:
  score = round(0.5 * content_accuracy_pct + 0.5 * structure_accuracy_pct)
(postprocess.recompute_calibrated_score 참조)

## SYSTEM

```
You are a meticulous evaluator of table extraction quality, using a strict,
decomposed scoring rubric. You will be shown the original image of a
table and the raw HTML extracted from it. In past evaluations, scores
clustered too high (most extractions scored 90+) even when many cells
or structural elements were actually wrong. Assume errors exist unless
you can positively verify otherwise, and use the FULL 0-100 range —
do not default to high scores. Respond ONLY with a single JSON object
matching the required schema.
```

## USER (템플릿, {candidate_html}은 추출된 HTML로 치환)

```
Extracted HTML (raw code):
{candidate_html}

Step 1 — Cell-level content accuracy: Compare every data cell (not
headers) in the image to the extracted HTML. Estimate
content_accuracy_pct: the percentage of data cells whose text content
is completely and exactly correct (integer 0-100).

Step 2 — Structural accuracy: Compare row/column layout, merged cells
(rowspan/colspan), and header hierarchy between the image and the HTML.
Estimate structure_accuracy_pct: the percentage of structural elements
(rows, columns, merges, header levels) that are correctly reproduced
(integer 0-100).

Step 3 — Enumerate up to 5 of the most significant errors (empty list
if none), each tagged with severity.

Calibration anchors (your percentages should behave like these):
- 100%: every cell/structural element verified correct.
- ~90%: 1-2 minor mismatches out of many elements.
- ~70%: a handful of cells/structural elements wrong, most correct.
- ~50%: roughly half right, half wrong or unverifiable.
- ~25%: only a small fraction correct.
- 0%: unrelated, empty, or not a table.

If you are unsure whether a cell or structural element is correct,
count it as INCORRECT (do not round up generously).

Return ONLY this JSON object:
{
  "content_accuracy_pct": <integer 0-100>,
  "structure_accuracy_pct": <integer 0-100>,
  "errors": [
    {"description": "<specific error>", "severity": "critical|major|minor"}
  ]
}
```

## JSON Schema

```json
{
  "type": "object",
  "properties": {
    "content_accuracy_pct": {
      "type": "integer"
    },
    "structure_accuracy_pct": {
      "type": "integer"
    },
    "errors": {
      "type": "array",
      "items": {
        "type": "object",
        "properties": {
          "description": {
            "type": "string"
          },
          "severity": {
            "type": "string",
            "enum": [
              "critical",
              "major",
              "minor"
            ]
          }
        },
        "required": [
          "description",
          "severity"
        ],
        "additionalProperties": false
      }
    }
  },
  "required": [
    "content_accuracy_pct",
    "structure_accuracy_pct",
    "errors"
  ],
  "additionalProperties": false
}
```
\end{Verbatim}
\paragraph{JSON schema.}
\begin{Verbatim}
# judge_calibrated_v2

이 파일은 `scripts/looprunner/prompts.py`의 스냅샷입니다. 실제 실행 시 소스는 prompts.py이며, 이 파일은 재현성/버전 추적을 위한 커밋용 기록입니다.

## Score 계산 방식

모델은 score를 반환하지 않는다. 코드가 결정적으로 재계산:
  score = round(0.5 * content_accuracy_pct + 0.5 * structure_accuracy_pct)
(postprocess.recompute_calibrated_score 참조)

## SYSTEM

```
You are a meticulous evaluator of table extraction quality, using a strict,
decomposed scoring rubric. You will be shown the original image of a
table and the raw HTML extracted from it. In past evaluations, scores
clustered too high (most extractions scored 90+) even when many cells
or structural elements were actually wrong. Assume errors exist unless
you can positively verify otherwise, and use the FULL 0-100 range —
do not default to high scores. Respond ONLY with a single JSON object
matching the required schema.
```

## USER (템플릿, {candidate_html}은 추출된 HTML로 치환)

```
Extracted HTML (raw code):
{candidate_html}

Step 1 — Cell-level content accuracy: Compare every data cell (not
headers) in the image to the extracted HTML. Estimate
content_accuracy_pct: the percentage of data cells whose text content
is completely and exactly correct (integer 0-100).

Step 2 — Structural accuracy: Compare row/column layout, merged cells
(rowspan/colspan), and header hierarchy between the image and the HTML.
Estimate structure_accuracy_pct: the percentage of structural elements
(rows, columns, merges, header levels) that are correctly reproduced
(integer 0-100).

Step 3 — Enumerate up to 5 of the most significant errors (empty list
if none), each tagged with severity.

Calibration anchors (your percentages should behave like these):
- 100%: every cell/structural element verified correct.
- ~90%: 1-2 minor mismatches out of many elements.
- ~70%: a handful of cells/structural elements wrong, most correct.
- ~50%: roughly half right, half wrong or unverifiable.
- ~25%: only a small fraction correct.
- 0%: unrelated, empty, or not a table.

If you are unsure whether a cell or structural element is correct,
count it as INCORRECT (do not round up generously).

Return ONLY this JSON object:
{
  "content_accuracy_pct": <integer 0-100>,
  "structure_accuracy_pct": <integer 0-100>,
  "errors": [
    {"description": "<specific error>", "severity": "critical|major|minor"}
  ]
}
```

## JSON Schema

```json
{
  "type": "object",
  "properties": {
    "content_accuracy_pct": {
      "type": "integer"
    },
    "structure_accuracy_pct": {
      "type": "integer"
    },
    "errors": {
      "type": "array",
      "items": {
        "type": "object",
        "properties": {
          "description": {
            "type": "string"
          },
          "severity": {
            "type": "string",
            "enum": [
              "critical",
              "major",
              "minor"
            ]
          }
        },
        "required": [
          "description",
          "severity"
        ],
        "additionalProperties": false
      }
    }
  },
  "required": [
    "content_accuracy_pct",
    "structure_accuracy_pct",
    "errors"
  ],
  "additionalProperties": false
}
```
\end{Verbatim}

\subsection{Pairwise \rid{judge\_pairwise\_v1} prompt}

\paragraph{System prompt.}
\begin{Verbatim}
# judge_pairwise_v1

이 파일은 `scripts/looprunner/prompts.py`의 스냅샷입니다. 실제 실행 시 소스는 prompts.py이며, 이 파일은 재현성/버전 추적을 위한 커밋용 기록입니다.

## 설계 의도

judge_v1(pointwise, 0-100 점수)의 평가 기준 서술(시맨틱 보존, 구조, key test)을 최대한 그대로 재사용하고, **점수 산출 부분만 A/B 선택으로 교체**한 pairwise 버전. 목적: Landesberg et al. (arXiv:2603.12520)의 "pairwise 완화"가 폐루프 후보 풀(저장된 8회차 출력)에서도 성립하는지 확인. 재생성 없음, judge 콜만.

- 후보 2개(A/B)를 표 이미지와 함께 제시, "이미지를 더 충실히 재현한 쪽"을 A/B로만 답하게 함 (JSON 스키마 강제, tie 불허).
- 위치 편향 통제: 각 비교를 (A,B)와 (B,A) 순서 모두 실행, 2콜의 승자가 일치할 때만 유효 비교. 불일치율은 position-inconsistency로 별도 보고.
- 모델/온도는 judge_v1과 동일: gemini-3.1-flash-lite, temperature 0.

## Choice 산출 방식

모델이 JSON의 `choice` 필드로 직접 반환 ("A" 또는 "B"). tie 옵션 없음 (강제 선택).

## SYSTEM

```
You are a meticulous evaluator of table extraction quality. You will be
shown the original image of a table and two candidate HTML extractions
of it, labeled A and B. Judge each candidate by considering how it would
render as a table. Do not evaluate stylistic HTML code quality; evaluate
whether the rendered table implied by the HTML preserves the content,
structure, and cell-to-header associations of the original image.
Respond ONLY with a single JSON object matching the required schema.
```

## USER (템플릿, {candidate_a}/{candidate_b}는 후보 HTML로 치환)

```
Candidate A (raw HTML code):
{candidate_a}

Candidate B (raw HTML code):
{candidate_b}

Evaluate each candidate against the table shown in the image.
Judge semantic preservation of values, headers, and their associations,
as well as the table structure (rows, columns, merged cells, header
hierarchy).

Apply this key test to each candidate: Could a reader who sees ONLY the
rendered form of the extracted HTML — without access to the original
image — unambiguously reconstruct every cell-to-header mapping and all
content of the original table? A candidate that fails this test in more
places is the less faithful one.

Decide which candidate more faithfully reproduces the content,
structure, and cell-to-header associations of the original table. You
must pick exactly one winner, even if the two candidates are very close
or the difference is small.

Return ONLY this JSON object:
{
  "choice": "A" or "B"
}
```

## JSON Schema

```json
{
  "type": "object",
  "properties": {
    "choice": {
      "type": "string",
      "enum": ["A", "B"]
    }
  },
  "required": ["choice"],
  "additionalProperties": false
}
```

## judge_v1 대비 변경점 요약

| 항목 | judge_v1 | judge_pairwise_v1 |
|---|---|---|
| 입력 | 이미지 + 후보 1개 | 이미지 + 후보 2개 (A/B) |
| 평가 기준 서술 (시맨틱 보존/구조/key test) | 원문 | 동일 문구를 두 후보 비교형으로 최소 수정 |
| 오류 열거 (up to 5 errors) | 있음 | 제거 (선택만 요구) |
| 출력 | errors[] + score 0-100 | choice "A"/"B" (tie 불허) |
| 모델/온도 | gemini-3.1-flash-lite, temp 0 | 동일 |
\end{Verbatim}
\paragraph{User prompt template.}
\begin{Verbatim}
# judge_pairwise_v1

이 파일은 `scripts/looprunner/prompts.py`의 스냅샷입니다. 실제 실행 시 소스는 prompts.py이며, 이 파일은 재현성/버전 추적을 위한 커밋용 기록입니다.

## 설계 의도

judge_v1(pointwise, 0-100 점수)의 평가 기준 서술(시맨틱 보존, 구조, key test)을 최대한 그대로 재사용하고, **점수 산출 부분만 A/B 선택으로 교체**한 pairwise 버전. 목적: Landesberg et al. (arXiv:2603.12520)의 "pairwise 완화"가 폐루프 후보 풀(저장된 8회차 출력)에서도 성립하는지 확인. 재생성 없음, judge 콜만.

- 후보 2개(A/B)를 표 이미지와 함께 제시, "이미지를 더 충실히 재현한 쪽"을 A/B로만 답하게 함 (JSON 스키마 강제, tie 불허).
- 위치 편향 통제: 각 비교를 (A,B)와 (B,A) 순서 모두 실행, 2콜의 승자가 일치할 때만 유효 비교. 불일치율은 position-inconsistency로 별도 보고.
- 모델/온도는 judge_v1과 동일: gemini-3.1-flash-lite, temperature 0.

## Choice 산출 방식

모델이 JSON의 `choice` 필드로 직접 반환 ("A" 또는 "B"). tie 옵션 없음 (강제 선택).

## SYSTEM

```
You are a meticulous evaluator of table extraction quality. You will be
shown the original image of a table and two candidate HTML extractions
of it, labeled A and B. Judge each candidate by considering how it would
render as a table. Do not evaluate stylistic HTML code quality; evaluate
whether the rendered table implied by the HTML preserves the content,
structure, and cell-to-header associations of the original image.
Respond ONLY with a single JSON object matching the required schema.
```

## USER (템플릿, {candidate_a}/{candidate_b}는 후보 HTML로 치환)

```
Candidate A (raw HTML code):
{candidate_a}

Candidate B (raw HTML code):
{candidate_b}

Evaluate each candidate against the table shown in the image.
Judge semantic preservation of values, headers, and their associations,
as well as the table structure (rows, columns, merged cells, header
hierarchy).

Apply this key test to each candidate: Could a reader who sees ONLY the
rendered form of the extracted HTML — without access to the original
image — unambiguously reconstruct every cell-to-header mapping and all
content of the original table? A candidate that fails this test in more
places is the less faithful one.

Decide which candidate more faithfully reproduces the content,
structure, and cell-to-header associations of the original table. You
must pick exactly one winner, even if the two candidates are very close
or the difference is small.

Return ONLY this JSON object:
{
  "choice": "A" or "B"
}
```

## JSON Schema

```json
{
  "type": "object",
  "properties": {
    "choice": {
      "type": "string",
      "enum": ["A", "B"]
    }
  },
  "required": ["choice"],
  "additionalProperties": false
}
```

## judge_v1 대비 변경점 요약

| 항목 | judge_v1 | judge_pairwise_v1 |
|---|---|---|
| 입력 | 이미지 + 후보 1개 | 이미지 + 후보 2개 (A/B) |
| 평가 기준 서술 (시맨틱 보존/구조/key test) | 원문 | 동일 문구를 두 후보 비교형으로 최소 수정 |
| 오류 열거 (up to 5 errors) | 있음 | 제거 (선택만 요구) |
| 출력 | errors[] + score 0-100 | choice "A"/"B" (tie 불허) |
| 모델/온도 | gemini-3.1-flash-lite, temp 0 | 동일 |
\end{Verbatim}
\paragraph{JSON schema.}
\begin{Verbatim}
# judge_pairwise_v1

이 파일은 `scripts/looprunner/prompts.py`의 스냅샷입니다. 실제 실행 시 소스는 prompts.py이며, 이 파일은 재현성/버전 추적을 위한 커밋용 기록입니다.

## 설계 의도

judge_v1(pointwise, 0-100 점수)의 평가 기준 서술(시맨틱 보존, 구조, key test)을 최대한 그대로 재사용하고, **점수 산출 부분만 A/B 선택으로 교체**한 pairwise 버전. 목적: Landesberg et al. (arXiv:2603.12520)의 "pairwise 완화"가 폐루프 후보 풀(저장된 8회차 출력)에서도 성립하는지 확인. 재생성 없음, judge 콜만.

- 후보 2개(A/B)를 표 이미지와 함께 제시, "이미지를 더 충실히 재현한 쪽"을 A/B로만 답하게 함 (JSON 스키마 강제, tie 불허).
- 위치 편향 통제: 각 비교를 (A,B)와 (B,A) 순서 모두 실행, 2콜의 승자가 일치할 때만 유효 비교. 불일치율은 position-inconsistency로 별도 보고.
- 모델/온도는 judge_v1과 동일: gemini-3.1-flash-lite, temperature 0.

## Choice 산출 방식

모델이 JSON의 `choice` 필드로 직접 반환 ("A" 또는 "B"). tie 옵션 없음 (강제 선택).

## SYSTEM

```
You are a meticulous evaluator of table extraction quality. You will be
shown the original image of a table and two candidate HTML extractions
of it, labeled A and B. Judge each candidate by considering how it would
render as a table. Do not evaluate stylistic HTML code quality; evaluate
whether the rendered table implied by the HTML preserves the content,
structure, and cell-to-header associations of the original image.
Respond ONLY with a single JSON object matching the required schema.
```

## USER (템플릿, {candidate_a}/{candidate_b}는 후보 HTML로 치환)

```
Candidate A (raw HTML code):
{candidate_a}

Candidate B (raw HTML code):
{candidate_b}

Evaluate each candidate against the table shown in the image.
Judge semantic preservation of values, headers, and their associations,
as well as the table structure (rows, columns, merged cells, header
hierarchy).

Apply this key test to each candidate: Could a reader who sees ONLY the
rendered form of the extracted HTML — without access to the original
image — unambiguously reconstruct every cell-to-header mapping and all
content of the original table? A candidate that fails this test in more
places is the less faithful one.

Decide which candidate more faithfully reproduces the content,
structure, and cell-to-header associations of the original table. You
must pick exactly one winner, even if the two candidates are very close
or the difference is small.

Return ONLY this JSON object:
{
  "choice": "A" or "B"
}
```

## JSON Schema

```json
{
  "type": "object",
  "properties": {
    "choice": {
      "type": "string",
      "enum": ["A", "B"]
    }
  },
  "required": ["choice"],
  "additionalProperties": false
}
```

## judge_v1 대비 변경점 요약

| 항목 | judge_v1 | judge_pairwise_v1 |
|---|---|---|
| 입력 | 이미지 + 후보 1개 | 이미지 + 후보 2개 (A/B) |
| 평가 기준 서술 (시맨틱 보존/구조/key test) | 원문 | 동일 문구를 두 후보 비교형으로 최소 수정 |
| 오류 열거 (up to 5 errors) | 있음 | 제거 (선택만 요구) |
| 출력 | errors[] + score 0-100 | choice "A"/"B" (tie 불허) |
| 모델/온도 | gemini-3.1-flash-lite, temp 0 | 동일 |
\end{Verbatim}

\subsection{\rid{gen\_cp\_v1} structure-preserving generation prompt}

\paragraph{System prompt.}
\begin{Verbatim}
# gen_cp_* 프롬프트 전문 스냅샷 (copy-preserving 2조건, 2026-07-08)

생성 시스템 프롬프트(GEN_SYSTEM)와 iter0(GEN_V1)은 main과 동일.

## GEN_SYSTEM (공통)

```
You are an expert at converting table images into faithful HTML tables.
```

## CP_PRESERVE_CONSTRAINTS (보존 제약, 두 조건 공통)

```
Preservation constraints (IMPORTANT):
- Treat the previous HTML as the default answer: keep it unchanged unless
  you have clear visual evidence from the image that a specific change is
  required.
- Do NOT needlessly change the number of rows or columns, the
  rowspan/colspan structure, or the placement of empty cells.
- Only make minimal edits that fix a clear omission or clear
  misrecognition; preserve everything else exactly as it is.
- If you find nothing that clearly needs fixing, output the previous HTML
  verbatim.
```

## gen_cp_nofeedback_v1 (Run F, iter>=1)

```
A previous attempt at converting this table image to HTML is shown
below. Carefully compare it against the table image.

Requirements:
- Reproduce ALL cell contents exactly as shown, including numbers,
  symbols, and formatting-relevant text.
- Reproduce the table structure faithfully: use rowspan/colspan for
  merged cells, and preserve header rows/hierarchy using <thead> and
  <th> where appropriate.
- If a cell contains a run of leader dots/periods used for visual
  alignment (e.g. "Item . . . . . . . 123"), collapse the run to a
  single ellipsis "..." rather than reproducing every dot. Never repeat
  any character or token in a loop.
- Output ONLY the HTML table, starting with <table> and ending with
  </table>. No markdown, no explanation, no code fences.

Preservation constraints (IMPORTANT):
- Treat the previous HTML as the default answer: keep it unchanged unless
  you have clear visual evidence from the image that a specific change is
  required.
- Do NOT needlessly change the number of rows or columns, the
  rowspan/colspan structure, or the placement of empty cells.
- Only make minimal edits that fix a clear omission or clear
  misrecognition; preserve everything else exactly as it is.
- If you find nothing that clearly needs fixing, output the previous HTML
  verbatim.

Previous attempt:
{previous_html}

Output ONLY the HTML table.
```

## gen_cp_feedback_v1 (Run G, iter>=1)

```
Convert this table image into a complete HTML table.

Requirements:
- Reproduce ALL cell contents exactly as shown, including numbers,
  symbols, and formatting-relevant text.
- Reproduce the table structure faithfully: use rowspan/colspan for
  merged cells, and preserve header rows/hierarchy using <thead> and
  <th> where appropriate.
- If a cell contains a run of leader dots/periods used for visual
  alignment (e.g. "Item . . . . . . . 123"), collapse the run to a
  single ellipsis "..." rather than reproducing every dot. Never repeat
  any character or token in a loop.
- Output ONLY the HTML table, starting with <table> and ending with
  </table>. No markdown, no explanation, no code fences.

A previous attempt is shown below, along with reviewer feedback.
Produce an improved version that fixes the identified issues while
preserving everything that was already correct.

Preservation constraints (IMPORTANT):
- Treat the previous HTML as the default answer: keep it unchanged unless
  you have clear visual evidence from the image that a specific change is
  required.
- Do NOT needlessly change the number of rows or columns, the
  rowspan/colspan structure, or the placement of empty cells.
- Only make minimal edits that fix a clear omission or clear
  misrecognition; preserve everything else exactly as it is.
- If you find nothing that clearly needs fixing, output the previous HTML
  verbatim.

Previous attempt:
{previous_html}

Reviewer feedback (issues found):
{judge_errors_formatted}

Output ONLY the HTML table.
```
\end{Verbatim}
\paragraph{Structure-preservation constraints.}
\VerbatimInput[
  fontsize=\scriptsize,
  breaklines=true,
  breakanywhere=true,
  firstline=14,
  lastline=23
]{gen_cp_v1.md}
\paragraph{No-feedback template.}
\VerbatimInput[
  fontsize=\scriptsize,
  breaklines=true,
  breakanywhere=true,
  firstline=29,
  lastline=60
]{gen_cp_v1.md}
\paragraph{Feedback template.}
\VerbatimInput[
  fontsize=\scriptsize,
  breaklines=true,
  breakanywhere=true,
  firstline=65,
  lastline=102
]{gen_cp_v1.md}

\subsection{Resolved model versions}

\begin{table*}[t]
  \centering
  \small
  \caption{Model versions resolved from the execution logs.}
  \label{tab:app-model-versions}
  \begin{tabular}{lll}
    \toprule
    Role & Configuration & Resolved version string \\
    \midrule
    generator / self-judge & \judgev, \calibv &
      \texttt{google/gemini-3.1-flash-lite-20260507} \\
    cross-judge & \crossnano &
      \texttt{openai/gpt-5.4-nano-20260317} \\
    cross-judge & \crossgpt &
      \texttt{openai/gpt-5.4-20260305} \\
    cross-judge & \crossopus &
      \texttt{anthropic/claude-4.6-opus-20260205} \\
    \bottomrule
  \end{tabular}
\end{table*}


\newcommand{\appendixEF}{%

\section{Generator Trend Check}
\label{app:E}
\label{app:generator-trends}

We performed a 100-table trend check that held the judge and protocol fixed
while changing only the generator. This is a scope-limited analysis of the
relation between signs across generators and oracle headroom, not a full
replication.

\begin{table*}[t]
  \centering
  \small
  \caption{Net loop effect by generator.}
  \label{tab:app-e-generators}
  \begin{tabular}{lrrl}
    \toprule
    Generator & mean(final$-$iter0) & Wilcoxon $p$ & Decision \\
    \midrule
    gemini-3.1-flash-lite & $-0.0121$ & 0.04448 & significant \\
    gpt-5.4-mini & $+0.0020$ & 0.683 & not significant \\
    qwen3.7-plus & $-0.0104$ & 0.3135 & not significant \\
    claude-haiku-4.5 & $+0.0917$ & 1.961e-10 & significant \\
    minimax-m3 & $-0.0153$ & 0.1152 & not significant \\
    kimi-k2.6 & $-0.0638$ & 5.765e-08 & significant \\
    \bottomrule
  \end{tabular}
\end{table*}

\begin{table*}[t]
  \centering
  \small
  \caption{Oracle headroom by generator.}
  \label{tab:app-e-headroom}
  \begin{tabular}{lrrr}
    \toprule
    Generator & mean iter0 & mean oracle & headroom \\
    \midrule
    gemini-3.1-flash-lite & 0.7884 & 0.8150 & $+0.0266$ \\
    gpt-5.4-mini & 0.7174 & 0.7685 & $+0.0511$ \\
    qwen3.7-plus & 0.7830 & 0.8047 & $+0.0217$ \\
    claude-haiku-4.5 & 0.5675 & 0.6951 & $+0.1276$ \\
    minimax-m3 & 0.7764 & 0.8239 & $+0.0474$ \\
    kimi-k2.6 & 0.8559 & 0.8743 & $+0.0184$ \\
    \bottomrule
  \end{tabular}
\end{table*}

Across 6 points including main, Spearman correlation between oracle
headroom and the net effect of the loop was 0.771 ($n=6$).

\appassetlabeled{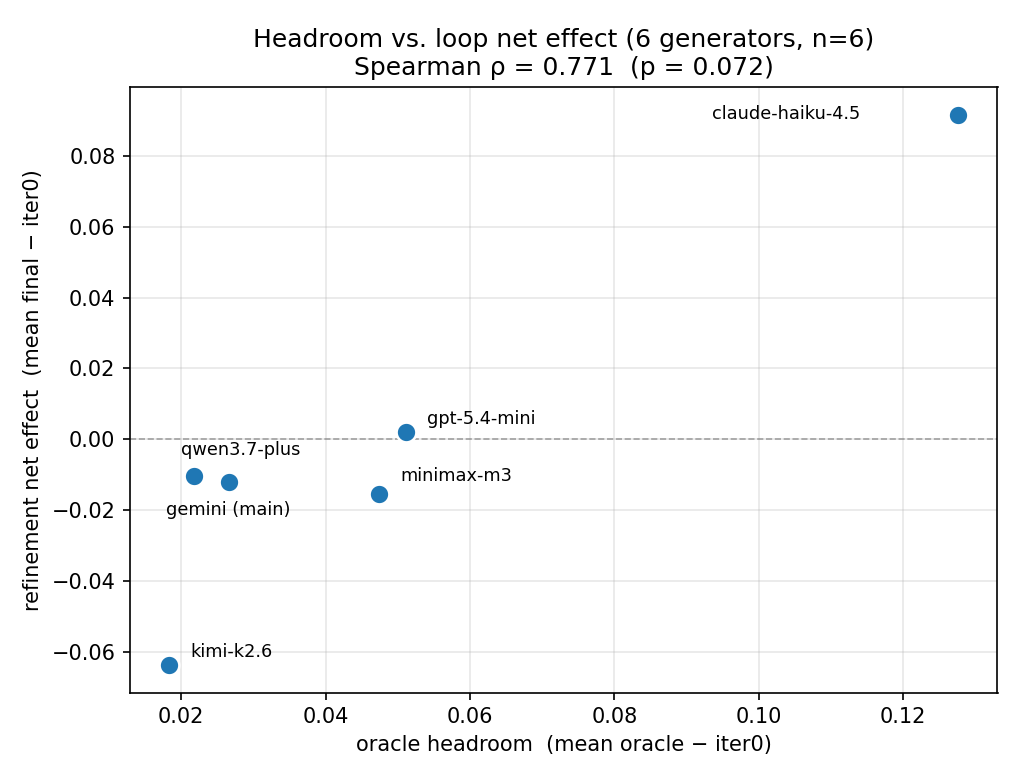}
  {\textbf{Oracle headroom versus the net effect of the loop (100-table means
  by generator, 6 points).} The judge and protocol are fixed and only the
  generator is changed. The horizontal axis is oracle minus iter0; the
  vertical axis is final minus iter0. The dotted line marks a net effect of
  0. Spearman correlation is 0.771 ($n=6$).}
  {fig:headroom}


\section{Additional Splits, Stratification, and Cost}
\label{app:F}
\label{app:additional-results}

\subsection{Sensitivity to including the development set}

The 20 development tables used in prompt design were excluded from the main
results. We also report results that include the development tables already
generated under the same conditions.

\begin{table*}[t]
  \centering
  \small
  \caption{Four-point comparison with and without the development set.}
  \label{tab:app-f-dev}
  \begin{tabular}{lrrrrr}
    \toprule
    Sample & $n$ & TEDS iter0 & TEDS final & \policy{best-by-judge} & oracle \\
    \midrule
    main excluding dev (480) & 476 & 0.7919 & 0.7718 & 0.7857 & 0.8104 \\
    including dev (500=480+dev20) & 496 & 0.7895 & 0.7707 & 0.7839 & 0.8102 \\
    \bottomrule
  \end{tabular}
\end{table*}

\subsection{Table-level outcome decomposition}

Because a mean can hide offsetting table-level outcomes, we jointly report
win/tie/loss counts, improvement and loss mass, the mean absolute change among
changed tables, and the severe-loss rate.

\begin{table*}[t]
  \centering
  \scriptsize
  \caption{Table-level decomposition of final$-$iter0 on FinTabNet and
    OmniDocBench.}
  \label{tab:app-f-decomposition}
  \resizebox{\textwidth}{!}{%
  \begin{tabular}{lrrrrrrrrr}
    \toprule
    Dataset & $n$ & win & tie & loss & win contribution & loss contribution &
      mean $\Delta$ & non-tie mean$|\Delta|$ & $\Delta<-0.10$ \\
    \midrule
    FinTabNet & 476 & 114 (23.9\%) & 164 (34.5\%) & 198 (41.6\%) &
      $+0.0117$ & $-0.0317$ & $-0.0200$ & 0.0663 & 56 (11.8\%) \\
    OmniDocBench & 272 & 56 (20.6\%) & 146 (53.7\%) & 70 (25.7\%) &
      $+0.0199$ & $-0.0179$ & $+0.0020$ & 0.0815 & 12 (4.4\%) \\
    \bottomrule
  \end{tabular}%
  }
\end{table*}

For the FinTabNet loss group, the mean was $-0.0763$ and the median was
$-0.0486$; for the win group, they were $+0.0489$ and $+0.0229$.
The means for the OmniDocBench win and loss groups were $+0.0965$ and
$-0.0694$.

\subsection{Stratified decomposition}

\begin{table*}[t]
  \centering
  \small
  \caption{OmniDocBench refinement results by complexity stratum.}
  \label{tab:app-f-strata}
  \begin{tabular}{lrrrrr}
    \toprule
    Stratum & $n$ & TEDS iter0 & $\Delta$ final$-$iter0 &
      selection gap & Spearman $\rho$ \\
    \midrule
    large & 22 & 0.8391 & $-0.0144$ & $+0.0102$ & 0.0898 \\
    simple & 154 & 0.7629 & $-0.0041$ & $+0.0194$ & 0.0227 \\
    span & 55 & 0.6805 & $+0.0268$ & $+0.0179$ & 0.0514 \\
    span\_large & 41 & 0.7771 & $+0.0005$ & $+0.0222$ & 0.1334 \\
    \bottomrule
  \end{tabular}
\end{table*}

FinTabNet selection stratification is reported in
Appendix~\ref{app:artifact-audit}. No source report was dedicated to
FinTabNet refinement win/tie/loss by stratum, so we do not newly aggregate
those values here.

\subsection{Measured cost and time}

The main run in Table~\ref{tab:app-f-cost} used 3,812 calls, cost \$13.0436,
and took 7.03 hours in serial execution.

\begin{table*}[t]
  \centering
  \small
  \caption{Measured API cost and serial runtime.}
  \label{tab:app-f-cost}
  \begin{tabular}{lrrr}
    \toprule
    Run & Calls & Cost (USD) & Serial time \\
    \midrule
    main generation (480$\times$8, gen+judge) & 3,812 & \$13.0436 & 7.03h \\
    \calibv{} re-scoring & 3,808 & \$3.7846 & 2.25h \\
    \crossnano{} re-scoring & 1,144 & \$0.8161 & 0.89h \\
    \crossgpt{} re-scoring & 600 & \$4.5235 & 0.46h \\
    \crossopus{} re-scoring & 600 & \$9.0049 & 0.91h \\
    \bottomrule
  \end{tabular}
\end{table*}

}

\newcommand{\appendixCD}{%
\section{Artifact Audit}
\label{app:C}
\label{app:artifact-audit}

\subsection{Correction for outputs containing multiple tables}

Because the stored TEDS implementation evaluates the first table in
multi-table HTML, we audited this artifact using an exclusion variant and a
merge-corrected variant.

Among 3,808 records, outputs containing at least 2 \texttt{<table>} tags
numbered 27 (0.71\%) and occurred in 9 tables.

\begin{table}[t]
  \centering
  \small
  \caption{Three result variants for the multi-table artifact.}
  \label{tab:app-c-multitable}
  \resizebox{\columnwidth}{!}{%
  \begin{tabular}{lrrr}
    \toprule
    Variant & $n$ & mean(final$-$iter0) & median \\
    \midrule
    reported & 476 & $-0.0200$ & $+0.0000$ \\
    multi-table excluded & 467 & $-0.0178$ & $+0.0000$ \\
    merge-corrected & 476 & $-0.0182$ & $+0.0000$ \\
    \bottomrule
  \end{tabular}%
  }
\end{table}

The difference in means between the merge-corrected and reported variants was
$+0.0018$. The sign of final$-$iter0 remained negative after correction.

\subsection{Audit across all tie-breaking rules}

When multiple iterations shared the highest judge score, we applied the
earliest, latest, and random rules to test the dependence of selection
conclusions on the rule.

\begin{table*}[t]
  \centering
  \scriptsize
  \caption{Complete audit of tie-breaking rules by judge.}
  \label{tab:app-c-tiebreak}
  \resizebox{\textwidth}{!}{%
  \begin{tabular}{lrrrrrr}
    \toprule
    Judge & $n$ & tables with ties & mean tied iterations &
      recovery (earliest) & recovery (latest) & recovery (random 100) \\
    \midrule
    \judgev & 476 & 90.3\% & 5.32 & $-33.2\%$ & $-96.4\%$ & $-73.4\%$ \\
    \calibv & 476 & 92.0\% & 5.82 & $-16.2\%$ & $-89.2\%$ & $-68.3\%$ \\
    \crossnano & 143 & 17.5\% & 1.23 & $-30.7\%$ & $-33.5\%$ & $-31.6\%$ \\
    \crossgpt & 75 & 26.7\% & 1.87 & $-43.4\%$ & $-35.1\%$ & $-41.2\%$ \\
    \crossopus & 75 & 69.3\% & 4.48 & $-19.6\%$ & $-22.8\%$ & $-23.9\%$ \\
    \bottomrule
  \end{tabular}%
  }
\end{table*}

Recovery for \judgev{} was $-173.3\%$ in large, 12.8\% in simple, 33.6\% in
span, and $-104.6\%$ in span\_large. Recovery for \calibv{} in the same order
was $-81.5\%$, $-21.4\%$, 13.5\%, and $-21.3\%$.


\section{Pairwise Evaluation Details}
\label{app:D}
\label{app:pairwise}

The pairwise tournament used calls in both directions with candidate order
reversed. Because neither candidate is more faithful when two candidates are
byte-identical, we excluded such pairs from conditional direction accuracy
and report only their frequency and positional inconsistency.

\begin{table*}[t]
  \centering
  \small
  \caption{Pairwise-selection recovery by dataset.}
  \label{tab:app-d-recovery}
  \resizebox{\textwidth}{!}{%
  \begin{tabular}{llrrr}
    \toprule
    Dataset & Strategy & $n$ & mean $\Delta$ & recovery \\
    \midrule
    FinTabNet & \policy{best-by-pairwise} & 476 & $-0.0191$ & $-102.7\%$ \\
    FinTabNet & \policy{random-among-8} & 476 & $-0.0142$ & $-76.3\%$ \\
    OmniDocBench & \policy{best-by-pairwise} & 271 & $+0.0060$ & 23.2\% \\
    OmniDocBench & \policy{random-among-8} & 271 & $+0.0027$ & 10.3\% \\
    \bottomrule
  \end{tabular}%
  }
\end{table*}

On FinTabNet, \policy{best-by-pairwise}$-$random was
$-0.0049\ [-0.0081,-0.0018]$, significantly worse than random. On
OmniDocBench, it was $+0.0034\ [+0.0000,+0.0069]$, significantly better than
random.

\subsection{Separating identical candidates and directional agreement}

Table~\ref{tab:app-d-identical} shows that identical candidates accounted for
53.6\% of FinTabNet and 63.8\% of OmniDocBench calls. Among distinct
candidates, agreement between the winner and the TEDS direction was 48.4\%
and 49.1\%, respectively.

\begin{table*}[t]
  \centering
  \small
  \caption{Identical-candidate share and conditional direction accuracy for
    pairwise calls.}
  \label{tab:app-d-identical}
  \resizebox{\textwidth}{!}{%
  \begin{tabular}{lrrrr}
    \toprule
    Dataset & identical share & distinct matches &
      distinct position inconsistency & winner-TEDS direction agreement \\
    \midrule
    FinTabNet & 53.6\% & 1,547 & 24.2\% & 48.4\% ($n=2,536$) \\
    OmniDocBench & 63.8\% & 687 & 30.1\% & 49.1\% ($n=1,117$) \\
    \bottomrule
  \end{tabular}%
  }
\end{table*}

\subsection{Candidate diversity and selection gain}

The mean number of unique candidates per table was 3.31 on FinTabNet and 2.71
on OmniDocBench, and the proportions of fully converged tables were 20.8\% and
39.1\%, respectively.

Spearman correlation between the number of unique candidates and the
best-minus-random gain was $-0.182\ [-0.277,-0.085]$ on FinTabNet and
$+0.178\ [+0.037,+0.329]$ on OmniDocBench. We did not observe a shared
direction in which greater diversity consistently increased selection gain.

\subsection{Seed repetitions}

Table~\ref{tab:app-d-seeds} shows that exact agreement across four seeds was
20.5\% for iteration index but 79.5\% for the hash of the selected HTML.

\begin{table}[t]
  \centering
  \small
  \caption{Pairwise-best repeatability across four seeds.}
  \label{tab:app-d-seeds}
  \resizebox{\columnwidth}{!}{%
  \begin{tabular}{lrl}
    \toprule
    Reproduction criterion & exact agreement & Interpretation \\
    \midrule
    iteration index & 8/39 = 20.5\% & lower bound \\
    best HTML hash & 31/39 = 79.5\% & substantive repeatability \\
    best TEDS & 33/39 = 84.6\% & metric-level repeatability \\
    \bottomrule
  \end{tabular}%
  }
\end{table}

}

OpenRouter routing was not fixed to a single backend. The two main runs used
both the ``Google AI Studio'' and ``Google'' backends, but the model version
string recorded in the logs was identical across calls.

\section{Verification Package}
\label{app:B}
\label{app:verification}

\subsection{V1: Cross-check against an independent TEDS implementation}

We cross-checked the stored metrics against
\rid{table\_recognition\_metric==0.0.6} from SWHL's independent
\rid{TableRecognitionMetric} repository.\footnote{\url{https://github.com/SWHL/TableRecognitionMetric}}
We applied both implementations to the same outputs.

Across 98 evaluation points, Pearson $r=0.99597$ ($p=2.42\mathrm{e}{-102}$)
and Spearman $\rho=0.99634$ ($p=2.25\mathrm{e}{-104}$). Absolute differences
had mean=0.00609, median=0.00310, p95=0.01827, and max=0.07340.
$|\Delta|>0.01$ occurred in 14/98 points and $|\Delta|>0.05$ in 1/98.
Recomputing the stored values with the same code produced absolute differences
of mean=0.000000 and max=0.000000. Differences between the independent
implementations arose from the normalization denominator when inline tags
were present, and the report records a PASS decision.

\subsection{V2: Repeatability of judge scoring}

Repeated identical scoring can have low intra-rater reliability
\citep{haldar2025ratingroulette}. We re-scored the same 39 tables, each with
8 iteration outputs, 3 times with \judgev{}.

All three scores agreed in 68.6\% of cases (214/312). Pairwise exact-agreement
rates across repetitions were 76.3\%, 81.1\%, and 79.8\%. For scores on the
same input, $|\max-\min|$ had mean=2.58, median=0.0, p95=10.0, and max=25.
Kendall's $W$ for within-table iteration rankings had mean=0.362 and
median=0.333; for $W<0.5$, the table proportion was 64.1\%. The set of highest-scoring
iterations was identical across all three repetitions for 41.0\% of tables
(16/39), while selection under the earliest-iteration rule agreed for 64.1\%
(25/39).

\begin{table}[t]
  \centering
  \small
  \caption{Recovery under the earliest-iteration rule by scoring repetition.}
  \label{tab:app-v2-recovery}
  \begin{tabular}{lrr}
    \toprule
    Score source & mean(best$-$iter0) & recovery \\
    \midrule
    original main scores & $-0.0088$ & $-42.2\%$ \\
    rep1 & $-0.0064$ & $-30.5\%$ \\
    rep2 & $-0.0012$ & $-5.5\%$ \\
    rep3 & $-0.0015$ & $-7.1\%$ \\
    \bottomrule
  \end{tabular}
\end{table}

\subsection{V3: Feedback-labeling protocol}

We classified feedback items in the stratified sample as REAL, PARTIAL,
GT-inconsistent, or UNVERIFIABLE. The analysis source data are preserved in
\texttt{feedback\_labels.csv} and \texttt{feedback\_sample40.csv}. The
\texttt{HALLUCINATED} column name in the source CSV corresponds to the
GT-inconsistent category in the paper and is preserved verbatim.

\begin{table}[t]
  \centering
  \small
  \caption{Feedback-audit label counts by stratum.}
  \label{tab:app-v3-labels}
  \resizebox{\columnwidth}{!}{%
  \begin{tabular}{lrrrrr}
    \toprule
    Stratum & REAL & PARTIAL & GT-inconsistent & UNVERIFIABLE & Total \\
    \midrule
    L1 & 1 & 0 & 12 & 0 & 13 \\
    L2 & 1 & 1 & 9 & 1 & 12 \\
    L3 & 2 & 1 & 5 & 2 & 10 \\
    L4 & 8 & 3 & 2 & 0 & 13 \\
    Overall & 12 & 5 & 28 & 3 & 48 \\
    \bottomrule
  \end{tabular}%
  }
\end{table}

Under the LLM-assisted labels, TEDS declined after 13/16 addressed
GT-inconsistent items (81.2\%, Wilson 95\% CI [57.0\%, 93.4\%]). TEDS
increased after 11/12 addressed REAL items (91.7\%, [64.6\%, 98.5\%]), and
12/13 L1 items were GT-inconsistent (92.3\%, [66.7\%, 98.6\%]). Recalculations
under the blinded author validation are reported in V3a.

\subsection{V3a: Blinded author validation and Pass 2 re-tagging}

The author validated the 48 feedback-audit labels in a separate blinded local
viewer. Top-level agreement was 39/48 = 81.2\%, increasing to 38/45 = 84.4\%
among high-confidence items, with Cohen's $\kappa=0.648$. The author-label
distribution was REAL 8, PARTIAL 5, GT-INCONSISTENT 34, and UNVERIFIABLE 1.

\begin{table*}[t]
  \centering
  \small
  \caption{Key recalculations comparing the LLM-assisted labels and blinded
    author validation.}
  \label{tab:app-author-validation}
  \begin{tabular}{lll}
    \toprule
    Metric & LLM labels & Author labels \\
    \midrule
    L1 GT-inconsistent proportion &
      12/13 (92.3\%; Wilson 66.7--98.6\%) &
      12/13 (92.3\%; Wilson 66.7--98.6\%) \\
    Decline after GT-inconsistent item addressed &
      13/16 (81.2\%; 57.0--93.4\%) &
      13/20 (65.0\%; 43.3--81.9\%) \\
    Increase after REAL item addressed &
      11/12 (91.7\%; 64.6--98.5\%) &
      7/8 (87.5\%; 52.9--97.8\%) \\
    \bottomrule
  \end{tabular}
\end{table*}

The 9 disagreements were VRSN.2006 item0
(UNVERIFIABLE$\to$PARTIAL), HOG.2014 item0
(UNVERIFIABLE$\to$GT-INCONSISTENT), KEY.2015 item0
(PARTIAL$\to$UNVERIFIABLE), ZBRA.2013 item0
(UNVERIFIABLE$\to$GT-INCONSISTENT), CHD.2012 item0
(REAL$\to$GT-INCONSISTENT), KMB.2010 item0
(REAL$\to$GT-INCONSISTENT), PEP.2017 item0
(PARTIAL$\to$GT-INCONSISTENT), PEP.2017 item1
(REAL$\to$PARTIAL), and PKI.2014 page 32 item0
(REAL$\to$GT-INCONSISTENT). The original claims for CHD.2012, KMB.2010, and
PKI.2014 page 32, all relabeled from REAL to GT-INCONSISTENT, assumed that the
leader dots or ellipses were absent from the image.

The 34 author-labeled GT-INCONSISTENT items were re-tagged in the calibrated
Pass 2 v2. The 2-button-plus-BOUNDARY scheme defined HALLUCINATION as cases in
which the target was not present in the image, CONVENTION as cases in which
the target was present but the prescription departed from the GT serialization,
and BOUNDARY as cases that could not be fixed confidently to either category.
The v1 tags were preserved but hidden in the v2 interface. The final
distribution was CONVENTION 28/34, HALLUCINATION 4/34, and BOUNDARY 2/34.
Restricting the analysis to the 28 items with an LLM subtype, the author v2
distribution was CONVENTION 25/28, HALLUCINATION 2/28, and BOUNDARY 1/28.
Among v1-to-v2 changes, of the 27 items projected as HALLUCINATION in v1,
24 moved to CONVENTION.

Pass 1 contained 4 GT-error-related notes and Pass 2 v2 contained 2 BOUNDARY
items, producing 5 unique candidates for the final GT-error/boundary record.
PKI.2014 page 32 item0 was tagged BOUNDARY in Pass 2 v2, with the note:
``Recheck the Pass 1 label: the premise of the claim is true and its
direction is toward GT; REAL/PARTIAL candidate.'' This item does not change
the final main-text numbers and is retained only as a recheck record.

\subsection{V4: IID independent-candidate contrast}

We used 8 independently generated candidates with no feedback chain to
separate candidate-pool contamination from the judge's ranking ability.

\begin{table*}[t]
  \centering
  \small
  \caption{\judgev{} selection versus a random baseline in IID best-of-8.}
  \label{tab:app-v4-iid}
  \begin{tabular}{lrrrrl}
    \toprule
    Tie rule & $n$ & best mean $\Delta$ & recovery &
      best$-$random [95\% CI] & Decision \\
    \midrule
    earliest & 40 & $+0.0053$ & 14.2\% &
      $+0.0027\ [-0.0138,+0.0184]$ & no difference \\
    latest & 40 & $+0.0023$ & 6.1\% &
      $-0.0003\ [-0.0176,+0.0145]$ & no difference \\
    random 100 & 40 & $+0.0071$ & 19.1\% &
      $+0.0045\ [-0.0065,+0.0136]$ & no difference \\
    \bottomrule
  \end{tabular}
\end{table*}

\subsection{V5: No-feedback control loop}

We omitted judge calls and injected only a fixed generic improvement
instruction, then paired the results with the same tables in the main
experiment.

\begin{table*}[t]
  \centering
  \small
  \caption{Paired results for the no-feedback control and main experiment.}
  \label{tab:app-v5-nofeedback}
  \begin{tabular}{lrrrrrr}
    \toprule
    Run & $n$ & mean $\Delta$ & Wilcoxon $p$ & win/tie/loss &
      severe loss & stagnation rate \\
    \midrule
    \ablnf & 99 & $-0.0122$ & 0.005371 & 17/48/34 &
      8 (8.1\%) & 83.7\% \\
    main & 99 & $-0.0122$ & 0.04448 & 25/34/40 &
      11 (11.1\%) & 54.1\% \\
    \bottomrule
  \end{tabular}
\end{table*}

The mean paired difference was $+0.0001$ (Wilcoxon signed-rank $p=0.7612$).

\appasset{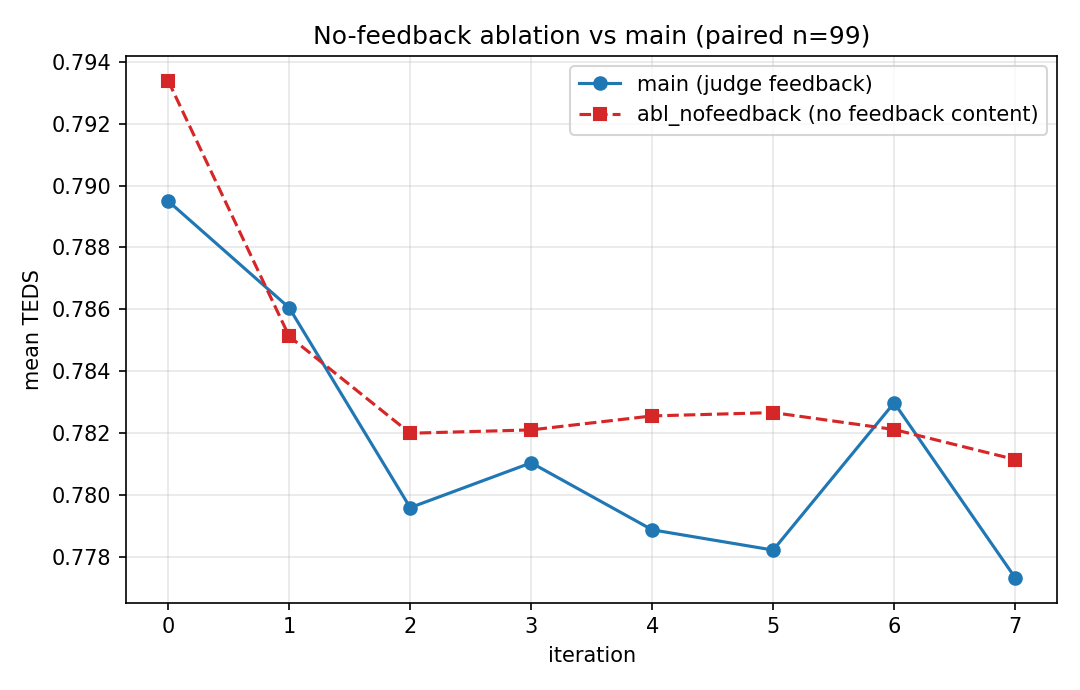}
  {Iteration curves for the no-feedback control.}

\subsection{V6: 4-condition preservation contrast}

We applied a paired $2\times2$ design crossing the presence of feedback with
the presence of a structure-preserving instruction on FinTabNet and
OmniDocBench. Table~\ref{tab:decomposition} in the main text reports the
contrast from the full-sample frozen-iter0 re-fork. The following two
tables give condition-level values from the original experiment, in which each
condition had its own iter0 and all four conditions completed a common set of
$n=99$ tables.

\begin{table*}[t]
  \centering
  \scriptsize
  \caption{Sensitivity analyses and bootstrap CIs for the full-n frozen-iter0
    B/C severe-loss-rate contrast.}
  \label{tab:app-v6-fulln-bc-ci}
  \resizebox{\textwidth}{!}{%
  \begin{tabular}{llrrrrr}
    \toprule
    Dataset & Handling & $n$ & B severe & C severe &
    C$-$B [bootstrap 95\% CI] & two-sided exact McNemar $p$ \\
    \midrule
    FinTabNet & carry-forward & 476 & 17/476 (3.6\%) & 4/476 (0.8\%) &
      $-2.7$ pp [$-4.4$, $-1.1$] & 0.0023 \\
    FinTabNet & complete-case & 450 & 16/450 (3.6\%) & 3/450 (0.7\%) &
      $-2.9$ pp [$-4.7$, $-1.3$] & 0.0010 \\
    OmniDocBench & carry-forward & 272 & 6/272 (2.2\%) & 2/272 (0.7\%) &
      $-1.5$ pp [$-2.9$, $-0.4$] & 0.1250 \\
    OmniDocBench & complete-case & 268 & 6/268 (2.2\%) & 2/268 (0.7\%) &
      $-1.5$ pp [$-3.0$, $-0.4$] & 0.1250 \\
    \bottomrule
  \end{tabular}%
  }
\end{table*}

\begin{table*}[t]
  \centering
  \small
  \caption{4-condition preservation contrast on FinTabNet.}
  \label{tab:app-v6-ft}
  \begin{tabular}{lrrrrrr}
    \toprule
    Condition & mean $\Delta$ & Wilcoxon $p$ & w/t/l &
      severe-loss rate & stagnation rate & full convergence \\
    \midrule
    \rid{main} & $-0.0122$ & 0.0445 & 25/34/40 & 11.1\% & 54.1\% & 20.2\% \\
    \ablnf & $-0.0122$ & 0.00537 & 17/48/34 & 8.1\% & 83.7\% & 39.4\% \\
    \cpnf & $-0.0006$ & 0.583 & 15/68/16 & 1.0\% & 88.7\% & 59.6\% \\
    \cpfb & $-0.0107$ & 0.133 & 27/31/41 & 12.1\% & 60.0\% & 18.2\% \\
    \bottomrule
  \end{tabular}
\end{table*}

The paired differences were main$-$abl $+0.0001$ ($p=0.761$),
\cpnf{}$-$\rid{abl} $+0.0116$ ($p=0.0225$), and
\cpfb{}$-$\rid{main} $+0.0015$ ($p=0.632$).

\begin{table*}[t]
  \centering
  \scriptsize
  \caption{4-condition preservation contrast on OmniDocBench.}
  \label{tab:app-v6-od}
  \begin{tabular}{lrrrrrrrr}
    \toprule
    Condition & mean $\Delta$ & $p$ & w/t/l & severe & loss mass &
      improvement mass & stagnation rate & structural-change rate \\
    \midrule
    \rid{main\_od} & $+0.0023$ & 0.75 & 19/57/23 & 3.0\% &
      $-1.750$ & $+1.982$ & 62.2\% & 23.2\% \\
    \rid{abl\_nofeedback\_od} & $+0.0014$ & 0.856 & 18/61/20 & 3.0\% &
      $-0.888$ & $+1.030$ & 83.3\% & 18.2\% \\
    \rid{cp\_nofeedback\_od} & $+0.0042$ & 0.681 & 15/68/16 & 0.0\% &
      $-0.304$ & $+0.716$ & 89.8\% & 15.2\% \\
    \rid{cp\_feedback\_od} & $+0.0179$ & 0.801 & 17/58/24 & 1.0\% &
      $-0.679$ & $+2.450$ & 70.1\% & 26.3\% \\
    \bottomrule
  \end{tabular}
\end{table*}

On OmniDocBench, \rid{main\_od}$-$\rid{abl\_od} was $+0.0009$
($p=0.911$), \rid{cp\_fb\_od}$-$\rid{cp\_nofb\_od} was $+0.0137$
($p=0.815$), \rid{cp\_nofb\_od}$-$\rid{abl\_od} was $+0.0027$
($p=0.589$), and \rid{cp\_fb\_od}$-$\rid{main\_od} was $+0.0155$
($p=0.328$).

\appasset{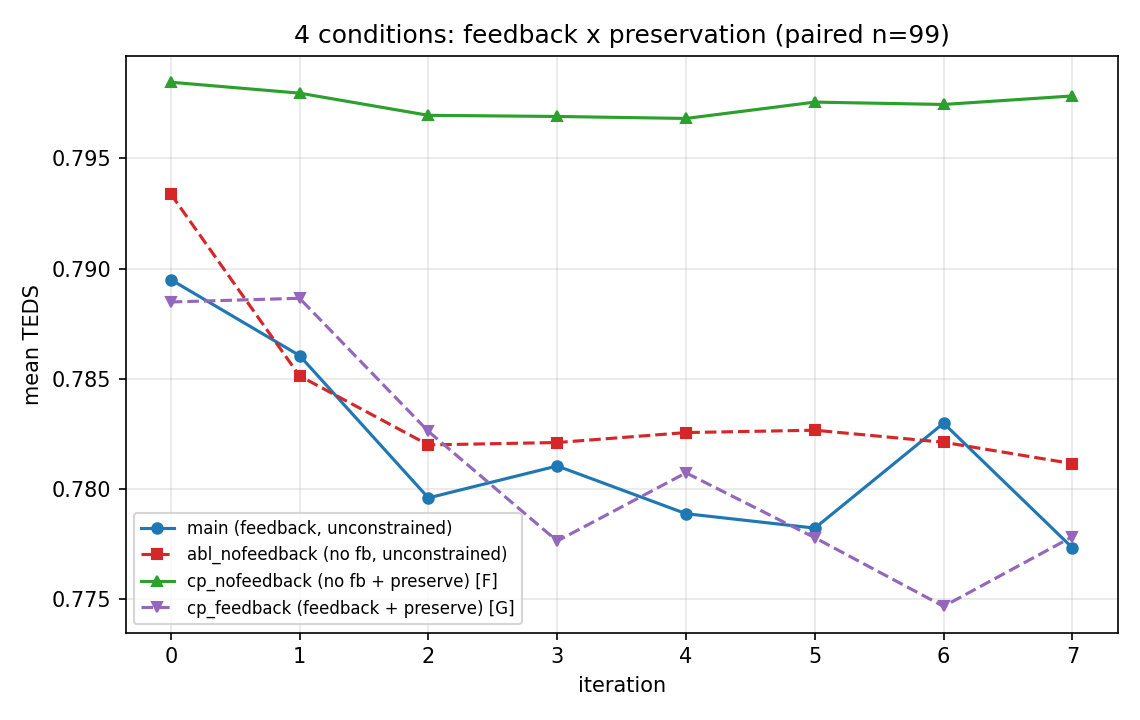}
  {Iteration curves for the 4 FinTabNet preservation conditions.}

\appasset{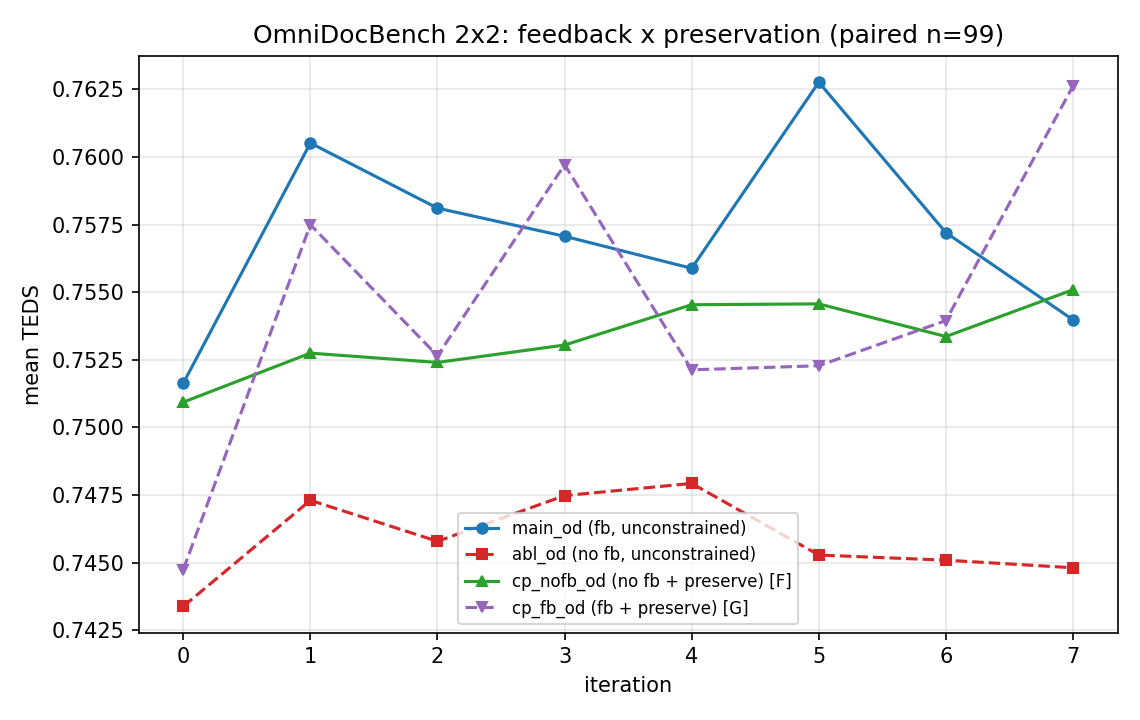}
  {Iteration curves for the 4 OmniDocBench preservation conditions.}

\subsection{V7: Decomposition of tie statistics}

The initial statistic $(N-U)/N$, computed from the total number of scores $N$
and the number of unique score values $U$, measures the coarseness of the score
alphabet rather than the actual probability of a candidate-pair tie. The
corrected statistics use unordered within-table candidate pairs as the
denominator.

\begin{table*}[t]
  \centering
  \scriptsize
  \caption{Comparison of score-alphabet coarseness and candidate-pair tie
    statistics.}
  \label{tab:app-v7-ties}
  \resizebox{\textwidth}{!}{%
  \begin{tabular}{llrrrrr}
    \toprule
    Dataset & Judge & $N;U$ & $(N-U)/N$ & all-pair ties &
      distinct-output tie rate & gap tie rate \\
    \midrule
    FinTabNet & \judgev & 3,806; 12 & 99.68\% & 65.95\% & 49.09\% & 42.24\% \\
    FinTabNet & \calibv & 3,804; 20 & 99.47\% & 68.91\% & 50.42\% & 44.64\% \\
    FinTabNet & \crossnano & 1,144; 47 & 95.89\% & 7.79\% & 6.42\% & 4.57\% \\
    FinTabNet & \crossgpt & 600; 43 & 92.83\% & 20.24\% & 12.74\% & 13.19\% \\
    FinTabNet & \crossopus & 600; 30 & 95.00\% & 56.29\% & 32.88\% & 25.46\% \\
    OmniDocBench & \judgev & 2,175; 16 & 99.26\% & 68.92\% & 37.50\% & 23.58\% \\
    \bottomrule
  \end{tabular}%
  }
\end{table*}

FinTabNet and OmniDocBench contained 6,358 and 4,540 byte-identical candidate
pairs, respectively. Their conditional distinct-output tie rates were 49.1\%
and 37.5\%. When restricted to distinct outputs with
$|\Delta\mathrm{TEDS}|>0.05$, the gap tie rates were 42.2\% and 23.6\%.

Recomputation using strict $t-1\rightarrow t$ transitions did not change the
reported direction-agreement or false-improvement rates for any configuration
at the displayed precision.

\begin{table*}[t]
  \centering
  \scriptsize
  \setlength{\tabcolsep}{2.7pt}
  \renewcommand{\arraystretch}{1.12}
  \caption{Judge diagnostic metrics for five judge configurations.
    Conditional direction accuracy and false-improvement rates use
    transitions not tied on either score or TEDS. Repeatability $W$ is
    Kendall's $W$ for iteration rankings from 3 scorings of the same input
    and was measured only for \judgev{} on FinTabNet. \textemdash{} indicates
    unverified.}
  \label{tab:app-judge-diagnostics}
  \resizebox{\textwidth}{!}{%
  \begin{tabular}{llrccccccc}
    \toprule
    Judge & Dataset & $n$ &
    \shortstack{Spearman $\rho$\\TEDS [95\% CI]} &
    \shortstack{Spearman $\rho$\\S-TEDS [95\% CI]} &
    \shortstack{Distinct\\score values} &
    \shortstack{Saturation\\$\geq95$ (\%)} &
    \shortstack{Conditional direction\\agreement} &
    \shortstack{False-improvement\\rate} &
    \shortstack{Repeatability\\Kendall's $W$} \\
    \midrule
    \judgev & FinTabNet & 476 &
      \shortstack{0.0956\\$[0.0177,0.1746]$} &
      \shortstack{0.0580\\$[-0.0192,0.1368]$} &
      12 & 79.82 & 0.6503 & 0.2000 & 0.362 \\
    \calibv & FinTabNet & 476 &
      \shortstack{0.1231\\$[0.0562,0.1910]$} &
      \shortstack{0.0926\\$[0.0244,0.1616]$} &
      20 & 89.20 & 0.7154 & 0.1671 & \textemdash \\
    \crossnano & FinTabNet & 143 &
      \shortstack{$-0.0639$\\$[-0.1738,0.0515]$} &
      \shortstack{$-0.0758$\\$[-0.1894,0.0379]$} &
      47 & 4.63 & 0.5539 & 0.2180 & \textemdash \\
    \crossgpt & FinTabNet & 75 &
      \shortstack{$-0.0299$\\$[-0.2149,0.1589]$} &
      \shortstack{$-0.0548$\\$[-0.2453,0.1514]$} &
      43 & 9.67 & 0.4520 & 0.3164 & \textemdash \\
    \crossopus & FinTabNet & 75 &
      \shortstack{$-0.0654$\\$[-0.2704,0.1541]$} &
      \shortstack{$-0.1341$\\$[-0.3341,0.0790]$} &
      30 & 48.83 & 0.5973 & 0.2215 & \textemdash \\
    \midrule
    \judgev & OmniDocBench & 272 &
      \shortstack{0.0294\\$[-0.0723,0.1353]$} &
      \shortstack{$-0.0175$\\$[-0.1222,0.0905]$} &
      16 & 78.80 & 0.4477 & 0.3073 & \textemdash \\
    \bottomrule
  \end{tabular}%
  }
\end{table*}


\begingroup
\sloppy
\appendixCD
\endgroup
\appendixEF

\section{Comparison with Self-Refine}
\label{app:G}
\label{app:self-refine}

Our loop shares the iterative feedback-and-revision pattern of Self-Refine, in
which the same model performs both operations. This study does not reproduce
the original paper. It tests a closed-loop configuration used in practice, in
which a reference-free judge supplies feedback and selection signals. We
specify the structural differences below.

\begin{table*}[t]
  \centering
  \scriptsize
  \caption{Structural comparison between Self-Refine Algorithm 1 and our loop.}
  \label{tab:app-g-selfrefine}
  \resizebox{\textwidth}{!}{%
  \begin{tabular}{rllll}
    \toprule
    \# & Component & Original Self-Refine & Our loop (main) & Difference \\
    \midrule
    1 & FEEDBACK &
      same model, few-shot, free-form text &
      same model in judge role, zero-shot JSON &
      natural language$\rightarrow$JSON; few-shot$\rightarrow$zero-shot \\
    2 & REFINE history &
      full history accumulated &
      preceding output and latest feedback only &
      no history accumulation \\
    3 & termination &
      stop indicator, maximum 4 iterations &
      fixed 8 iterations, no early stopping &
      no stop indicator \\
    4 & prompt format &
      few-shot, actionable/specific &
      zero-shot, JSON schema, preserve-correct instruction &
      few-shot$\rightarrow$zero-shot \\
    5 & self condition &
      same model generates, gives feedback, and revises &
      generator=judge=Gemini &
      same \\
    \bottomrule
  \end{tabular}%
  }
\end{table*}

The results therefore should not be interpreted as evidence against the
original Self-Refine implementation as a whole. The tested object is our
closed-loop configuration, which does not accumulate history and uses
zero-shot structured feedback and a fixed iteration count.


\section{Novelty Reanalysis}
\label{app:H}
\label{app:novelty}

For the 99 tables shared by the main and no-feedback conditions, we define
$D_i=\Delta_{\mathrm{main},i}-\Delta_{\mathrm{nofeedback},i}$. A value
$D_i>0$ indicates that feedback benefits that table, while $D_i<0$ indicates
harm.

Mean $D=+0.0001$, median $=+0.0000$, and Wilcoxon signed-rank $p=0.761$.
Win/tie/loss counts were 35/34/30. Mean $|D|$ was 0.0408, whereas
$|\mathrm{mean}\ D|$ was 0.0001. Spearman correlation between table-level
changes in the two conditions was 0.465 ($p=1.21\mathrm{e}{-06}$).

\appasset{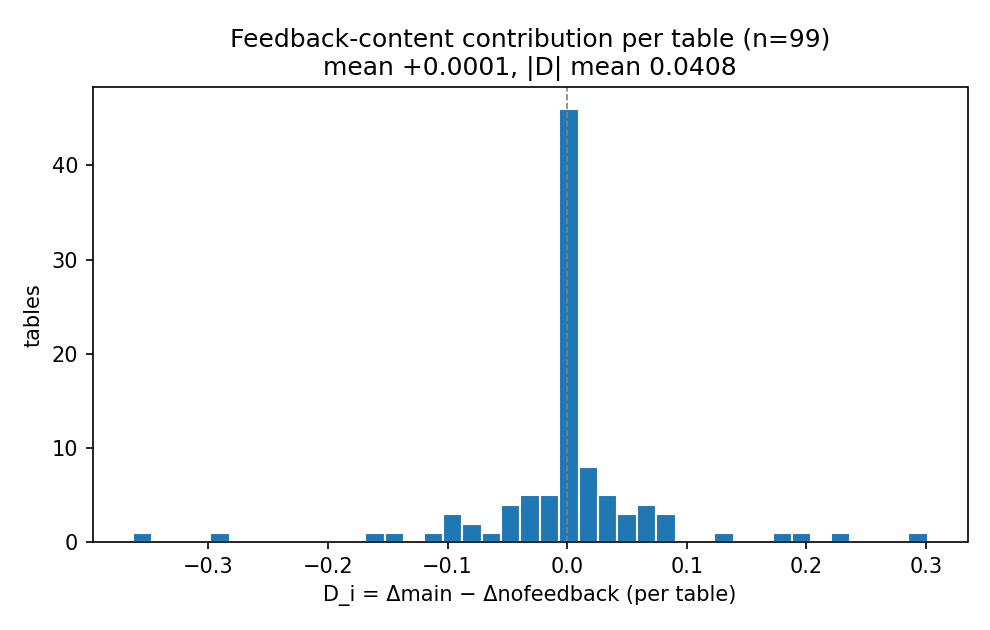}
  {Distribution of the table-level difference $D_i$ between the main and
  no-feedback conditions.}

\subsection{Decomposition by iter0 quality, headroom, and complexity}

\begin{table*}[t]
  \centering
  \small
  \caption{Decomposition of $D_i$ by iter0 TEDS quartile.}
  \label{tab:app-h-iter0}
  \begin{tabular}{lrrrr}
    \toprule
    Quartile & $n$ & Range & mean $D$ & w/t/l \\
    \midrule
    Q1 (low) & 25 & [0.359, 0.717] & $+0.0298$ & 11/10/4 \\
    Q2 & 25 & [0.718, 0.836] & $-0.0053$ & 8/9/8 \\
    Q3 & 24 & [0.842, 0.899] & $-0.0041$ & 7/9/8 \\
    Q4 (high) & 25 & [0.901, 0.955] & $-0.0203$ & 9/6/10 \\
    \bottomrule
  \end{tabular}
\end{table*}

Concentration of values collapsed the headroom decomposition into two groups.
Q1 contained 74 tables, range [0.000, 0.012], mean $D=-0.0122$, and
w/t/l=17/31/26. Q2 contained 25 tables, range [0.015, 0.441], mean
$D=+0.0363$, and w/t/l=18/3/4. We interpret these results only as descriptive
statistics.

\begin{table*}[t]
  \centering
  \small
  \caption{Decomposition of $D_i$ by complexity stratum.}
  \label{tab:app-h-strata}
  \begin{tabular}{lrrrrr}
    \toprule
    Stratum & $n$ & mean $D$ & w/t/l & mean $\Delta_{\rm main}$ &
      mean $\Delta_{\rm abl}$ \\
    \midrule
    large & 20 & $-0.0372$ & 6/4/10 & $-0.0512$ & $-0.0140$ \\
    simple & 16 & $+0.0155$ & 4/11/1 & $+0.0193$ & $+0.0038$ \\
    span & 31 & $+0.0247$ & 12/10/9 & $+0.0083$ & $-0.0165$ \\
    span\_large & 32 & $-0.0083$ & 13/9/10 & $-0.0233$ & $-0.0151$ \\
    \bottomrule
  \end{tabular}
\end{table*}

\subsection{Intersection with V3 labels}

The intersection between the V3 labeled sample and the common 99 tables
contained 8 tables. Of these, 6 tables containing a GT-inconsistent item had
mean $D=-0.0191$, while the other 2 had mean $D=-0.0115$. Because the denominators
are small, we report these as raw descriptive statistics without a directional
claim.

\subsection{Predictive association of churn and structural change with decline}

\begin{table*}[t]
  \centering
  \small
  \caption{Comparison of churn with and without feedback.}
  \label{tab:app-h-churn}
  \begin{tabular}{lrr}
    \toprule
    Metric & \rid{main} & \ablnf \\
    \midrule
    changed-transition rate & 45.9\% & 16.3\% \\
    mean $|\Delta|$ per transition & 0.0237 & 0.0063 \\
    mean $|\Delta|$ among changed transitions & 0.0517 & 0.0387 \\
    decline share among changed transitions & 43.1\% & 43.4\% \\
    standard deviation of table-level $\Delta$ & 0.0841 & 0.0653 \\
    mean gross movement $\sum_t|\Delta_t|$ & 0.1660 & 0.0441 \\
    \bottomrule
  \end{tabular}
\end{table*}

Both conditions had a net effect of $-0.0122$, and the difference between
their net effects was 0.0001.

\begin{table*}[t]
  \centering
  \small
  \caption{Transition-level association between structural change and decline
    in the main condition.}
  \label{tab:app-h-structure}
  \begin{tabular}{lr}
    \toprule
    Metric & Value \\
    \midrule
    $P(\text{decline}>0.05\mid\text{structural change})$ &
      26.9\% ($n=171$) \\
    $P(\text{decline}>0.05\mid\text{content-only change})$ &
      8.2\% ($n=147$) \\
    severe declines accompanied by structural change & 85.2\% (23/27) \\
    declines accompanied by structural change & 79.3\% (46/58) \\
    share of total decline from structural-change transitions &
      73.4\% (6.469/8.819) \\
    \bottomrule
  \end{tabular}
\end{table*}

These values are offline transition-level associations and do not reflect
changes in later states that would result from placing a guard inside the
actual loop. They should therefore be interpreted as an upper bound on
preventability.


\section{Additional Main-Text Details}
\label{app:I}

\subsection{Experimental details moved for the page limit}

\paragraph{Failure handling.}
API calls were retried up to 5 times with exponential backoff, although both
main runs required 0 actual retries. HTML was extracted from the first
\texttt{table} tag through the last closing tag using a regular expression.
Extraction failed for 1.0\% of main outputs and 1.3\% of OmniDocBench outputs.
When this occurred at iter0, the affected table was excluded from analysis
(4 main tables: 3 cases exceeding the token limit and 1 API error); when it
occurred at iteration 1 or later, the last successful output was carried
forward. When judge JSON parsing failed (0.2\% in main and 0.09\% in
OmniDocBench), the iteration proceeded with a generic improvement instruction
and no error list. No prompt correction or selective retry was applied.

\paragraph{Temperature and iteration budget.}
We chose temperature 0 to maintain a clean causal interpretation. Under
near-deterministic generation, the difference between iteration $t$ and
iteration $t+1$ is attributable mainly to feedback injected into the prompt.
Full determinism does not hold, as measured in Appendix~V2.\footnote{The same
model version was routed through two serving backends (log literals ``Google
AI Studio'' and ``Google''; main 1,952/1,860 calls, OmniDocBench 1,081/1,096
calls). We cannot rule out serving variability as a contributor to scoring
non-reproducibility at temperature 0.} A higher-temperature condition (0.2)
increased output diversity in pilot experiments but did not improve quality
(Appendix~\ref{app:B}). We adopted 8 iterations after frequent score ties made
judge-based early stopping effectively inoperative. Section~4.2 compares
fixed-$k$ termination policies.

\paragraph{Sampling details.}
For FinTabNet, we used the
\rid{docling-project/FinTabNet\_OTSL} test split \citep{lysak2023otsl},
downloaded on 2026/7/2, and
the original FinTabNet HTML in its \rid{html} field as GT. This differs from
the corrected and canonicalized FinTabNet.c \citep{smock2023aligning}. We
defined 4 mutually exclusive complexity strata using OTSL merge tokens and a
cell-count threshold of 104 (test-split p90): simple 80, span 160, large 100,
and span\_large 160. Of 500 sampled tables, 20 development tables (5 per
stratum) were excluded, leaving 480 (75/155/95/155); 4 iter0 failures yielded
$n=476$. Results including the 20 development tables are in
Appendix~\ref{app:F}. For OmniDocBench, we used the 1,651-page release
downloaded on 2026/7/2, which extends the 981-page release described in the
CVPR paper. Approximate stratification of 273 English or mixed English-Chinese
tables used HTML span attributes and a 100-cell threshold: simple 154, span
56, large 22, and span\_large 41. One iter0 failure yielded $n=272$.

\paragraph{Data licenses and terms.}
The original FinTabNet release is documented as CDLA-Permissive-1.0, whereas
the \rid{FinTabNet\_OTSL} dataset card labels the converted release's license
as ``other.''\footnote{\url{https://huggingface.co/datasets/docling-project/FinTabNet_OTSL}}
OmniDocBench's dataset copyright statement limits use to research and excludes
commercial use; its repository's Apache-2.0 license covers the evaluation
code, not necessarily the dataset.\footnote{\url{https://huggingface.co/datasets/opendatalab/OmniDocBench}}
We used both datasets only for evaluation, trained no model on them, and do
not redistribute either dataset or its source PDFs in the submission package.

\paragraph{Judge and metric details.}
The five configurations comprised the self configuration (476/272 tables), a
redesigned score-production configuration (476), another family at the same
tier (143), a higher-tier same-family model (75), and a third family (75). The
redesigned configuration was selected on 20 development tables. The
143-table sample came from a pre-specified 150 after excluding 6 overlaps and
1 failure; the two 75-table configurations share a fixed-seed subset. The 3
cross judges re-score stored outputs without re-running the loop.

TEDS is deterministic under fixed GT and implementation, but is not independent
of GT serialization and canonicalization conventions. We cross-validated our
implementation on 98 points (Pearson and Spearman 0.996;
Appendix~\ref{app:B}). It evaluates only the first table in an output, affecting
0.71\% of outputs; exclusion and merged-table correction preserve every
conclusion (final minus iter0 $=-0.0200/-0.0178/-0.0182$;
Appendix~\ref{app:C}). Selection policies were \policy{always-iter0},
\policy{best-by-judge} under 3 tie-breaking rules,
\policy{best-by-pairwise} with 7 matches and 2 reversed-order calls,
\policy{random-among-8} with 1,000 draws, \policy{always-final}, and
\policy{oracle-best}. Recovery rate is mean($\Delta$ vs.\ iter0) divided by
mean(oracle $-$ iter0). ``Robustly beats'' requires a table-cluster bootstrap
95\% CI supporting an advantage over random on both datasets and under all
three tie-breaking rules.

\subsection{Repeated-Measures Test: Cochran's Q}

In the re-forked $2\times2$ experiment, all 203 parsing failures were
truncations at the output-token limit. Re-testing the four conditions as
repeated measures on each table found differences in per-table any-failure on
FinTabNet (Cochran's $Q(3)=18.60$, $p=0.0003307$), but not on OmniDocBench
($Q(3)=6.65$, $p=0.084$). In six post hoc paired exact McNemar tests, only the
FinTabNet B-D comparison remained significant after Holm correction (raw
$p=0.00098$, Holm $p=0.0059$). The severe-loss contrast yielded the same
decision under complete-case and carry-forward handling, mitigating this
confound. Interactions involving feedback and structure preservation remain
entangled with condition-asymmetric parsing failures and are therefore
reported as exploratory only.

\subsection{Detailed selection diagnostics}
\label{app:I-selection}

\begingroup
\let\originalcaption\caption
\renewcommand{\caption}[1]{\originalcaption{\textbf{Judge discrimination and
selection diagnostics (5 judges, FinTabNet main; \judgev{} also reported for
OmniDocBench).} The distinct-output tie rate is the proportion of byte-level
different within-table pairs assigned the same score. The gap tie rate
restricts this to pairs separated by at least 0.05 TEDS. Conditional direction
accuracy is computed among score-distinguished pairs. Mean $\Delta$ and the
vs.-random verdict use the earliest-iteration tie rule.}}
\begin{table*}[t]
  \centering
  \small
  \setlength{\tabcolsep}{4.0pt}
  \renewcommand{\arraystretch}{1.10}
  \caption{}
  \label{tab:judge-diagnostics}
  \resizebox{\textwidth}{!}{%
  \begin{tabular}{llrccccl}
    \toprule
    Judge & Dataset & $n$ &
    \shortstack{Distinct-output\\tie rate} &
    \shortstack{Gap tie\\rate} &
    \shortstack{Conditional direction\\accuracy} &
    \shortstack{Mean $\Delta$\\vs.\ iter0} &
    vs.-random verdict \\
    \midrule
    \judgev   & FinTabNet & 476 & 49.1\% & 42.2\% & 61.5\% & $-0.0062$ &
      sig.\ better than random \\
    \calibv   & FinTabNet & 476 & 50.4\% & 44.6\% & 65.1\% & $-0.0030$ &
      sig.\ better than random \\
    \crossnano & FinTabNet & 143 & 6.4\% & 4.6\% & 54.8\% & $-0.0061$ &
      not disting.\ from random \\
    \crossgpt  & FinTabNet & 75 & 12.7\% & 13.2\% & 45.9\% & $-0.0102$ &
      not disting.\ from random \\
    \crossopus & FinTabNet & 75 & 32.9\% & 25.5\% & 57.9\% & $-0.0046$ &
      not disting.\ from random \\
    \midrule
    \judgev   & OmniDocBench & 272 & 37.5\% & 23.6\% & 47.2\% & $+0.0072$ &
      sig.\ better than random \\
    \bottomrule
  \end{tabular}%
  }
  \vspace{2pt}

  \begin{minipage}{0.99\textwidth}
    \footnotesize
    OmniDocBench cross-judge re-scoring was not run. Recovery percentages
    are omitted because they duplicate
    Table~\ref{tab:selection-policies}.
  \end{minipage}
\end{table*}

\endgroup

The self-judge and redesigned score-production configuration tied on 49.1\%
and 50.4\% of distinct-output pairs (66.0\% and 68.9\% of all pairs). They
also tied on 42.2\% and 44.6\% of pairs separated by at least 0.05 TEDS and
used only 12 to 20 score values. Yet the lowest-tie configurations (6.4\% and
12.7\%) still had low or negative TEDS rank correlation and did not recover
improvement. Rankings changed across three repeated scorings ($W=0.36$), and
score-quality correlation was at most 0.13 for all five judges. The failure
also held for independent candidates (Appendix~V4).

For \judgev{}, conditional direction accuracy was 61.5\% on FinTabNet
(Wilson 95\% CI [59.8\%, 63.2\%], treating pairs as independent) and 47.2\%
on OmniDocBench ([44.8\%, 49.6\%]). The values were at the 83.8th
($p=0.3257$) and 97.0th ($p=0.0619$) percentiles of a tie-block permutation
null. FinTabNet lay within the central 90\% range; OmniDocBench exceeded the
upper boundary, but its two-sided $p$-value was 0.0619. This null does not
identify a causal tie-breaking effect or the share of policy advantage due to
the tie rule.

Switching from pointwise scoring to pairwise comparison increased recovery
from 21\% to 61\% in prior one-step work
\citep{landesberg2026bestofn}, but did not transfer here. Pairwise selection
was significantly worse than pointwise and random on FinTabNet and was not
distinguishable from pointwise on OmniDocBench. Direction accuracy was 48 to
49\%. All 8 candidates were identical for 21 to 39\% of tables, but pairwise
still did not beat random among tables with at least 4 unique candidates.

\subsection{Detailed feedback-content audit}

\begingroup
\let\originalcaption\caption
\renewcommand{\caption}[1]{\originalcaption{\textbf{Feedback-content labeling
(40 stratified tables, 48 items).} Claims about iter0 were classified against
GT. Strata are based on $\Delta$(final $-$ iter0).}}
\begin{table*}[t]
  \centering
  \scriptsize
  \setlength{\tabcolsep}{3.0pt}
  \renewcommand{\arraystretch}{1.10}
  \caption{}
  \label{tab:feedback-labeling}
  \resizebox{\columnwidth}{!}{%
  \begin{tabular}{lrrrrrrrr}
    \toprule
    Stratum & Real & Partial & \shortstack{GT-\\inconsistent} &
    \shortstack{Unveri-\\fiable} & Total & Real (\%) &
    \shortstack{Addressed\\items} &
    \shortstack{Down among\\addressed (\%)} \\
    \midrule
    L1 (large loss) & 1 & 0 & 12 & 0 & 13 & 7.7 & 9 & 66.7 \\
    L2              & 1 & 1 &  9 & 1 & 12 & 8.3 & 7 & 85.7 \\
    L3              & 2 & 1 &  5 & 2 & 10 & 20.0 & 4 & 50.0 \\
    L4 (improvement)& 8 & 3 &  2 & 0 & 13 & 61.5 & 12 & 0.0 \\
    \midrule
    Overall         & 12 & 5 & 28 & 3 & 48 & 25.0 & 32 & 43.8 \\
    \bottomrule
  \end{tabular}%
  }
  \vspace{2pt}

  \begin{minipage}{0.98\columnwidth}
    \footnotesize
    Strata are defined by $\Delta(\text{final}-\text{iter0})$ over 40 sampled
    tables. Five tables had no error items; the remaining 35 tables yielded
    48 labeled items. The final column is the share of addressed items whose
    iter0-to-iter1 table-level TEDS direction was down.
  \end{minipage}
\end{table*}

\endgroup

Labeling used an LLM-assisted procedure. In blinded author validation,
top-level agreement was 39/48 = 81.2\% (38/45 = 84.4\% for high-confidence
items; Cohen's $\kappa=0.648$). Under author labels, the L1 GT-inconsistent
rate remained 12/13 = 92.3\%, while decline after addressing a GT-inconsistent
claim weakened to 13/20 = 65.0\% [43.3\%, 81.9\%]. The main table therefore
uses the original labels. L1 claims were 92\% GT-inconsistent (12/13, Wilson
95\% CI [67\%, 99\%]), while 62\% of L4 claims were real. Convention-type
claims were the majority (18/28). In the author's Pass 2 v2 re-tagging, the 34
GT-inconsistent items comprised 28 CONVENTION, 4 HALLUCINATION, and 2 BOUNDARY;
among 28 items with an original subtype, 25/28 were CONVENTION. Under the
LLM-assisted labels, 81\% of addressed GT-inconsistent claims co-occurred with
decline (13/16, [57\%, 93\%]), while 92\% of addressed real claims co-occurred
with improvement (11/12, [65\%, 99\%]). These small-sample associations should
not be read as causal.

Most of the 28 GT-inconsistent items reflect conflict between semantic HTML
conventions and GT serialization. S-TEDS also fell in these examples, so the
damage was structural. Severe breakage occurred in 11.8\% of FinTabNet and
4.4\% of OmniDocBench tables. Breakage also occurred after a score of 100 and
0 reported errors, and real claims could cause breakage when revision was
poorly executed.

\FloatBarrier
\subsection{Iteration curves and random-relative recovery}

\ifdefined\ARXIVFINAL\else
\begin{figure*}[t]
  \centering
  \includegraphics[width=\textwidth]{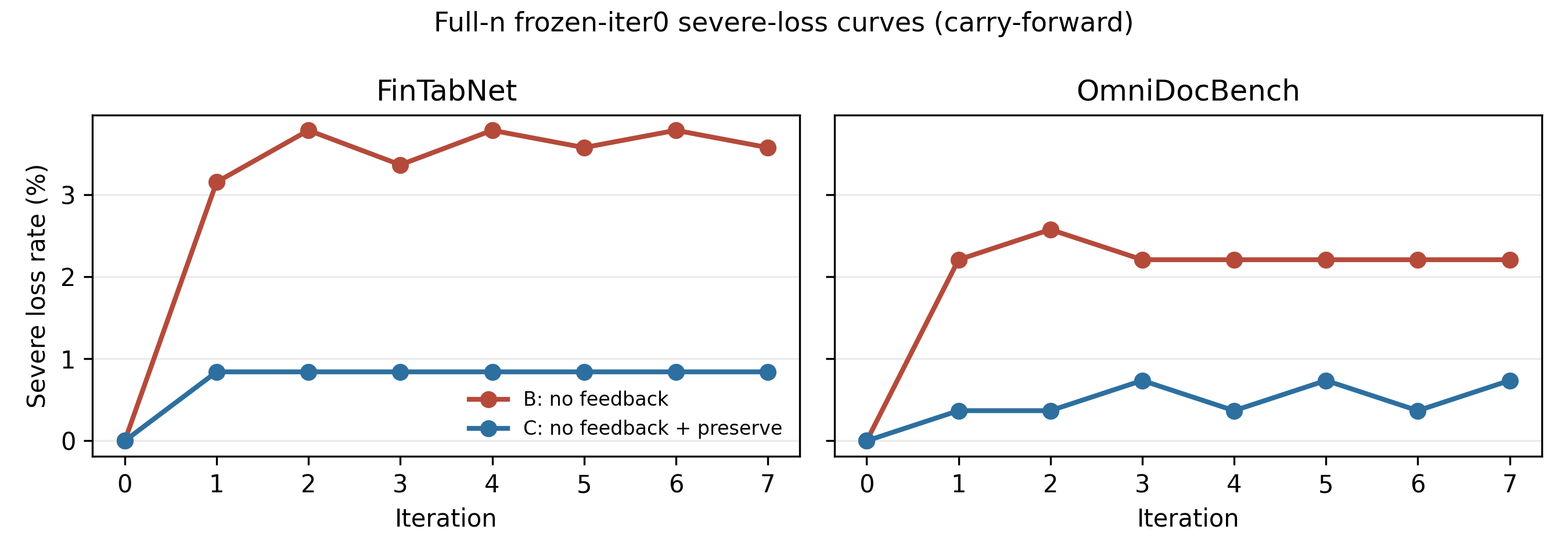}
  \caption{\textbf{Severe-loss rate by iteration in the full-n frozen-iter0
  B/C re-fork.} The final rate on FinTabNet fell from 3.6\% to 0.8\%
  (C$-$B = $-2.7$ pp, two-sided exact McNemar $p=0.0023$);
  OmniDocBench was directionally consistent, from 2.2\% to 0.7\%
  ($p=0.1250$).}
  \label{fig:conditions}
\end{figure*}
\fi

\ifdefined\ARXIVFINAL\else
\begin{figure*}[t]
  \centering
  \includegraphics[width=\textwidth]{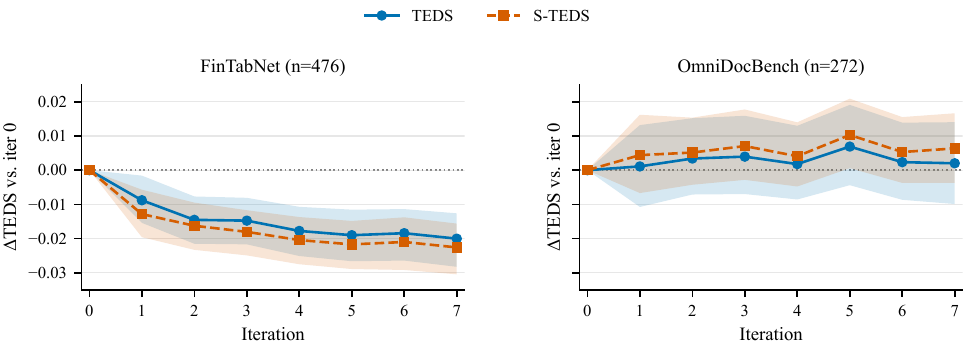}
  \caption{\textbf{Mean TEDS and S-TEDS by iteration on both datasets.}
  Values are changes relative to iter0. Shading gives table-cluster bootstrap
  95\% CIs.}
  \label{fig:iteration-curves}
\end{figure*}
\fi

Prior work reports approximately 21\% random-relative oracle recovery, defined
as $(E[O_{\mathrm{judge}}]-E[O_{\mathrm{random}}])/
(E[O_{\mathrm{oracle}}]-E[O_{\mathrm{random}}])$. Under the same
ratio-of-means definition, \judgev{} recovered 24.5\% on FinTabNet (bootstrap
95\% CI [12.6\%, 36.6\%]) and 19.4\% on OmniDocBench ([0.4\%, 37.5\%]).
These values do not change the negative or weak iter0-relative net effect.

\subsection{Expanded practical implications}

The sign of the loop's net effect is difficult to predict because opportunities
for conflict between GT conventions and judge preferences are difficult to
measure without labeled data. A small labeled audit should precede deployment.
Even a neutral mean concealed individual breakage in 4 to 12\% of tables, and
some broken outputs retained perfect judge scores. Retaining the first output
was therefore the safest baseline, although low-quality generators with large
headroom showed a conditional indication of benefit.

Level 1 is to stop feedback injection and add a prompt-level
structure-preservation constraint. In the paired 99-table comparison, mean
degradation was $-0.0006$ and not significant. Severe-loss reduction was
statistically significant in the full-sample FinTabNet re-fork and
directionally consistent on OmniDocBench, but exploratory results did not
show stable protection when feedback was retained. Level 2 is a deterministic structural guard for changes in
merge structure and row or cell counts. TEN, PICARD, and SynCode provide
related precedents \citep{mehrotra2026ten,scholak2021picard,ugare2025syncode},
although grammatical validity does not guarantee agreement with content, merge
structure, or GT conventions. Level 3 is not to use judge scores as a gating
signal. Among 12 addressed real claims, 11 co-occurred with improvement, which
suggests that verified feedback can be useful, but claim verification alone
does not prevent net degradation. The conditional large-table signal from
\calibv{} is not confirmed given its sample and selection procedure.


\endgroup

\section{Guarded Acceptance Policies}
\label{app:guarded}

Both guarded policies are simulated post hoc on the logged candidate stream
of the main loop (Section~3.4). The margin rule scans iterations in order,
starting from iter0 as the accepted output, and replaces the accepted output
only when a candidate's judge score exceeds the accepted output's score by
at least $\delta$; the 3 of 5{,}984 candidates whose judge score failed to
parse cannot be accepted. The stop-on-clean rule returns the first iteration
whose judge report contains no errors. A rollback-to-best-score variant
coincides with judge-max selection by construction and is not reported
separately. The grid $\delta\in\{0,20,40,60,80,100\}$ spans the observed
judge score scale. Confidence intervals are paired table-level bootstrap
with 10{,}000 resamples. Recovery follows the definition in Section~3.4.
Excluding the three tables containing an unparsed judge score changes mean
TEDS by at most 0.0005 for every policy and changes no conclusion.

\begin{table*}[t]
  \centering
  \small
  \setlength{\tabcolsep}{5pt}
  \caption{Guarded acceptance on the logged candidate stream. $\Delta$ is
    mean TEDS minus iter0 with paired bootstrap 95\% CI; Repl.\ is the
    number of tables whose final output differs from iter0.}
  \label{tab:guarded-grid}
  \begin{tabular}{llrrrr}
    \toprule
    Dataset & Policy & Mean TEDS & $\Delta$ vs.\ iter0 & Recovery & Repl. \\
    \midrule
    FinTabNet & $\delta=0$   & 0.7739 & $-0.0179$ $[-0.0250, -0.0108]$ & $-96.4\%$ & 302 \\
    FinTabNet & $\delta=20$  & 0.7899 & $-0.0020$ $[-0.0059, +0.0016]$ & $-10.8\%$ & 12 \\
    FinTabNet & $\delta=40$  & 0.7906 & $-0.0013$ $[-0.0039, +0.0011]$ & $-6.9\%$  & 5 \\
    FinTabNet & $\delta\in\{60,80,100\}$ & 0.7919 & $+0.0000$ $[+0.0000, +0.0000]$ & $0.0\%$ & 0 \\
    FinTabNet & stop-on-clean & 0.7755 & $-0.0164$ $[-0.0234, -0.0096]$ & $-88.3\%$ & 282 \\
    \midrule
    OmniDocBench & $\delta=0$   & 0.7598 & $+0.0052$ $[-0.0059, +0.0166]$ & $+20.2\%$ & 124 \\
    OmniDocBench & $\delta=20$  & 0.7661 & $+0.0116$ $[+0.0032, +0.0214]$ & $+44.5\%$ & 27 \\
    OmniDocBench & $\delta=40$  & 0.7639 & $+0.0093$ $[+0.0030, +0.0172]$ & $+35.8\%$ & 15 \\
    OmniDocBench & $\delta=60$  & 0.7546 & $+0.0001$ $[+0.0000, +0.0001]$ & $+0.2\%$  & 2 \\
    OmniDocBench & $\delta\in\{80,100\}$ & 0.7545 & $+0.0000$ $[+0.0000, +0.0000]$ & $0.0\%$ & 0 \\
    OmniDocBench & stop-on-clean & 0.7604 & $+0.0059$ $[-0.0049, +0.0171]$ & $+22.7\%$ & 100 \\
    \bottomrule
  \end{tabular}
\end{table*}

\end{document}